\pdfoutput=1
\documentclass[11pt]{article}

\usepackage{acl}

\usepackage{times}
\usepackage{latexsym}
\usepackage{amsmath}

\usepackage{graphicx}
\usepackage{bbold}
\usepackage{multirow}
\usepackage{pifont}
\usepackage{float}
\usepackage{lipsum} 
\usepackage{placeins}
\usepackage{booktabs}
\usepackage{calrsfs}
\usepackage{todonotes}
\usepackage{caption}
\usepackage{subcaption}
\usepackage{cleveref}
\usepackage{CJKutf8}
\usepackage{hyperref}
\newcommand{\zh}[1]{\begin{CJK}{UTF8}{gbsn}#1\end{CJK}}
\usepackage{float}
\usepackage{stfloats} 

\usepackage[T1]{fontenc}
\usepackage[utf8]{inputenc}

\usepackage{microtype}

\usepackage{inconsolata}

\newcommand*{\affaddr}[1]{#1} % No op here. Customize it for different styles.
\newcommand*{\affmark}[1][*]{\textsuperscript{#1}}
\newcommand*{\email}[1]{\texttt{#1}}
\newcommand\tab[1][0.5cm]{\hspace*{#1}}

\title{Towards Zero-Shot Multimodal Machine Translation}

\author{%
Matthieu Futeral\affmark[1,2]~~~Cordelia Schmid\affmark[1,2]~~~\textbf{Benoît Sagot}\affmark[1]~~~\textbf{Rachel Bawden}\affmark[1] \\
\affaddr{\affmark[1]Inria, Paris}\\
\affaddr{\affmark[2]Département d’informatique de l’ENS, CNRS, PSL Research University}\\
\email{firstname.lastname@inria.fr}
}

\begin{document}
\maketitle
\begin{abstract}
%Current multimodal machine translation (MMT) systems rely on fully supervised data whereby models are trained on sentences with their translations and accompanying images. However, this type of data is costly to collect and it limits the extension of MMT to other language pairs for which such data does not exist. In this work, we propose a method to bypass the need for fully supervised data to train MMT systems. We show that we can train MMT models using multimodal English data only and obtain disambiguation level close to state-of-the-art MMT models trained additionally on fully supervised examples. To prove that our method generalizes to languages without fully supervised data, we extend the recently release CoMMuTE evaluation dataset to three new languages: Arabic, Russian and Chinese. We further show that we can control the trade-off between disambiguation capabilities and translation fidelity at inference time using classifier-free guidance and without any additional data. Our code, data and trained models will be made publicly accessible.

Current multimodal machine translation (MMT) systems rely on fully supervised data (i.e~sentences with their translations and accompanying images), which is costly to collect and prevents the extension of MMT to language pairs with no such data. We propose a method to bypass the need for fully supervised data to train MMT systems, using multimodal English data only. 
%We first translate the data with a strong text-only machine translation (MT) model. 
%We then adapt the MT model to images by freezing it and introducing trainable lightweight modules. We next train it with (1) visually-conditioned masked language modelling (VMLM) on English captions and their images to learn to exploit the visual inputs and (2) Kullback-Leibler divergence on the synthetic translations and the accompanied images to keep general translation capabilities close to the original MT system.
Our method (\mbox{ZeroMMT}) consists in adapting a strong text-only machine translation (MT) model by training it jointly on two objectives: visually conditioned masked language modelling and the Kullback-Leibler divergence between the original MT and new MMT outputs.
We evaluate on standard MMT benchmarks and on CoMMuTE, a contrastive test set designed to evaluate how well models use images to disambiguate translations. ZeroMMT obtains disambiguation results close to state-of-the-art MMT models trained on fully supervised examples. To prove that ZeroMMT generalizes to languages with no fully supervised training data, we extend CoMMuTE to three new languages: Arabic, Russian and Chinese. We also show that we can control the trade-off between disambiguation capabilities and translation fidelity at inference time using classifier-free guidance and without any additional data. Our code, data and trained models are publicly accessible.\footnote{\url{https://github.com/MatthieuFP/CoMMuTE}}$^,$\footnote{\url{https://github.com/MatthieuFP/zerommt}}
\end{abstract}

\section{Introduction}

% add somewhere that traditional MMT systems only have +1 BLEU point from text-only baseline which is not significant at all. (add the ref) => in the Related Work
% add the extension of CoMMuTE to new languages

Multimodal machine translation (MMT) refers to the use of additional modalities, such as images or videos, in machine translation (MT) systems. The main purpose is to provide an additional signal in the case of ambiguity in the text to be  translated (i.e~the text alone does not provide enough information). Most current MMT models are trained solely on the Multi30K (M30K) dataset \citep{elliott-etal-2016-multi30k, elliott-EtAl:2017:WMT, barrault-etal-2018-findings}, a multilingual and multimodal corpus composed of 30K images, their English captions and translations in French, German and Czech. There have been recent breakthroughs in MMT thanks to the use of pretrained text-only MT systems and monolingual captioning data in order to adapt MT systems to MMT \citep{futeral-etal-2023-tackling, gupta2023cliptrans, vijayan2024adding}. Good results have been shown using this strategy on CoMMuTE \citep{futeral-etal-2023-tackling}, a benchmark designed to evaluate MMT models on their use of images to disambiguate between contrastive translations, and these results were significantly better than MMT systems trained on M30K only \citep{yin-etal-2020-novel, yao-wan-2020-multimodal, liu2021gumbel, wu-etal-2021-good, Li_2022_CVPR}. However, these models still rely on the multilingual {\em and} multimodal M30K corpus during training to ensure good translation performance. This presents a core limitation: collecting translations of captions is costly,\footnote{Authors of M30K stated they spent €23,000 on the translation of the 30,000 English captions into German.} restricting MMT's extension to new languages. Zero-shot transfer between languages has been tested to bypass the problem \citep{hirasawa-etal-2023-visual}, but this results in the poor exploitation of the visual modality to disambiguate ambiguous texts. %(i.e.~to correctly exploit the visual modality).

In this work, we address this limitation by proposing a method requiring only monolingual multimodal text data (i.e. English text-image pairs), removing the need for fully supervised data, i.e.~parallel and multimodal data such as M30K. We start from a strong pretrained MT system and use it to translate multimodal English data into the target languages of interest. We then adapt the pretrained MT system to images using two objectives: (1)~visually conditioned masked language modelling (VMLM) \citep{li-etal-2019-visualbert,lu-etal-2019-vilbert} on multimodal English data to force the model to use image information and (2)~a KL penalty on the translated multimodal data to maintain translation capabilities. We test our method on six languages directions: English to French, Czech, German, Arabic, Russian and Chinese, extending the CoMMuTE dataset to cover the three additional languages. Our method, called ZeroMMT, obtains CoMMuTE scores close to the supervised state of the art, while there is only a small drop in BLEU and COMET scores compared to the underlying text-only MT system on standard MMT benchmarks composed mainly of unambiguous examples (i.e.~where images are not useful for correct translation). %All of this without including any fully supervised data during training. 
We further show that we can control the trade-off between disambiguation and general translation performance at inference time with classifier-free guidance. %We eventually extend CoMMuTE to three additional language directions including two left behind the MMT community: English$\rightarrow$\{Chinese,Russian,Arabic\}.

\section{Related Work}

\paragraph{Training MMT systems}
% Good for misconceived reasons => regularisation
% no good training data => in fully supervised setting model can learn to ignore image to translate correctly
Research in MMT originally focused on which visual features to use \citep{li-etal-2022-vision} and how to integrate them into sequence-to-sequence models \citep{sutskever2014sequence} trained from scratch on the widely used M30K benchmark \citep{libovicky-etal-2016-cuni, calixto-etal-2016-dcu, elliott-kadar-2017-imagination, calixto-liu-2017-incorporating, yin-etal-2020-novel, liu2021gumbel, Li_2022_CVPR}. These MMT systems typically show improvements of around 1-2 BLEU points on standard MMT benchmarks in comparison to text-only baselines trained from scratch, which is not significant enough to state that MMT systems are better than their text-only counterparts \citep{mathur-etal-2020-tangled}. \citet{wu-etal-2021-good} observed that while they obtained +1 BLEU on average on M30K test sets with the use of images, they got the same improvements with randomly initialized visual features, most likely due to regularization, i.e.~the images were in reality not being exploited effectively. On top of that, being trained from scratch on fully supervised MMT data only, these models lag far behind state-of-the-art MT systems \citep{costa2022no} trained on large amounts of parallel text.

\citet{futeral-etal-2023-tackling} show that M30K contains few ambiguous examples requiring visual context, and that models can get good results on the benchmark while still struggling to exploit images correctly. They introduce VGAMT, an adapted MMT model based on a frozen state-of-the-art MT model. They also show that visually masked language modelling (VMLM) on English captioning data was a key additional objective to force MMT systems to become truly multimodal. \citet{sato-etal-2023-choosing} and \citet{bowen2024detecting} further show that choosing the masked tokens in a smart way instead of randomly slightly boosts results. However, these methods still require fully supervised data to be good at translation; training on VMLM alone results in a collapse in translation capabilities.

A few works have used pseudo-multimodal parallel data by translating English captions into the target language using a pretrained MT system \citep{li-etal-2021-vision, caglayan-etal-2021-cross, vijayan2024adding}. However, \citet{caglayan-etal-2021-cross} and \citet{vijayan2024adding} used them in a pretraining step in a form of distillation of the knowledge of the MT system into the new MMT model before fine-tuning on M30K. \citet{li-etal-2021-vision} use backtranslation to translate English captions into Turkish to train a Turkish-to-English MMT model for disambiguating gender pronouns from Turkish to English. While effective, their method cannot be applied beyond this particular context because it requires the MT system to output the correct translation, which cannot be assumed to be the case in more general ambiguous contexts (i.e.~when text context is not enough to translate the English text correctly).

There have been efforts to train MMT models without using fully supervised data \citep{su2019unsupervised,huang-etal-2020-unsupervised-multimodal,fei-etal-2023-scene}% in an unsupervised manner
. These approaches are however fundamentally different from this work as their goal is to obtain MT models using synthetic text-only parallel data through the use of visual pivoting, not targeting disambiguation capabilities. \citet{hirasawa-etal-2023-visual} proposed a zero-shot method to learn MMT by training on the little fully supervised data available aiming for zero-shot cross-lingual transfer. As the amount of fully supervised data for a single language is small ($\leq$ 30K text-image pairs), and few languages are covered ($\leq$ 8), this method results in poor exploitation of the image to learn disambiguation capabilities.

\begin{figure*}[!hbtp]
    \centering
    \includegraphics[width=.87\textwidth]{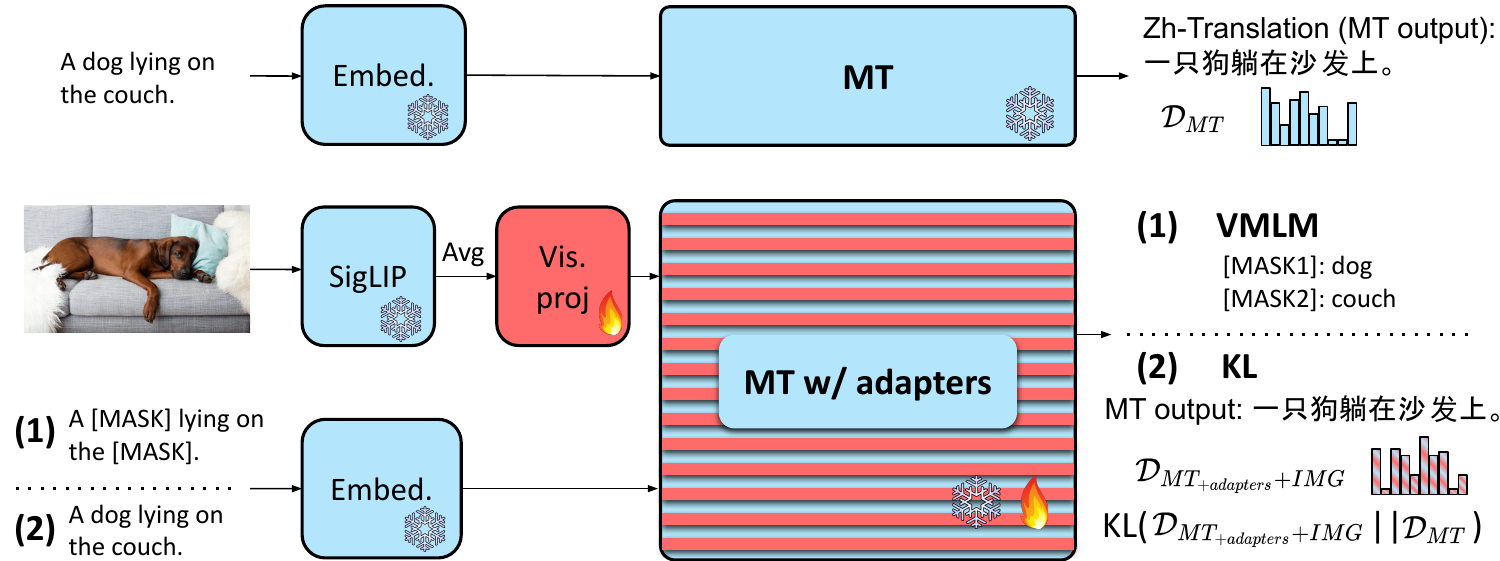}
    \caption{Overview of our approach. We train on two objectives: Visually conditioned masked language modelling (VMLM) and Kullback-Leibler (KL) divergence. All weights are frozen during training except the visual projector and the adapters in the MT model.}
    \label{fig:main-fig}
\end{figure*}

\paragraph{Evaluating MMT systems}
The test sets typically used to evaluate MMT systems are the test subsets of M30K \citep{elliott-etal-2016-multi30k, elliott-EtAl:2017:WMT, barrault-etal-2018-findings}. However, some of the translations were produced without access to the images and they have also been found to contain only a few ambiguous examples where visual context is necessary \citep{futeral-etal-2023-tackling}. They are therefore not best adapted to evaluating MMT systems. \citet{elliott-2018-adversarial} and \citet{caglayan-etal-2019-probing} proposed to use an adversarial evaluation method and a probing method based on masked text inputs to assess the utility of images in translation. However, this is not a good proxy for evaluating an MMT model's capacity to disambiguate translations as it relies on masked inputs and so says little about models' capacities to disambiguate when given unmasked text inputs. \citet{lala-specia-2018-multimodal}, \citet{li-etal-2021-vision} and \citet{zhu-etal-2023-beyond} released evaluation datasets composed of sentences in English with ambiguous words accompanied with disambiguating images. However, \citet{li-etal-2021-vision} only target the ambiguity of gender pronouns, and all these datasets are prone to distributional bias, which is difficult to measure and is such that text-only MT systems can perform very well on them (i.e.~images are in fact often not necessary for correct translation). Traditional MT metrics \citep{papineni-etal-2002-bleu, banerjee-lavie-2005-meteor, rei-etal-2020-comet} are also unable to catch how well MMT systems use images as they do not specifically target translations where images would be required to translate correctly. Tackling these issues, \citet{futeral-etal-2023-tackling} introduced CoMMuTE, a contrastive evaluation dataset, composed of English sentences to be translated, built around ambiguous words, each sentence accompanied by two translations with two images, each of which disambiguates the English sentence. 
MMT models are evaluated on their capacity to give a lower perplexity to the correct translation than the incorrect one, given the source sentence and an image. As perplexity is used to evaluate MMT models, it is a direct proxy of MMT models' capacity to disambiguate English sentences. Furthermore, text-only MT systems can only perform as well as random (50\%), as they do not have access to the images.

\section{Extending the CoMMuTE benchmark} \label{ss:commute}

\begin{table}[ht]
    \centering\small
    \resizebox{\linewidth}{!}{\begin{tabular}{lcccc} \toprule
         & \hphantom{ }En & \hphantom{ }Ar & \hphantom{ }Ru & \hphantom{ }Zh \\ \midrule
       \#unique sents. & \hphantom{1,}155 & \hphantom{1,}310 & \hphantom{1,}310 & \hphantom{1,}310 \\
       \#tokens & 1,384 & 2,958 & 3,105 & 2,832 \\
       \#unique toks. & \hphantom{1,}559 & \hphantom{1,}870 & 1,002 & \hphantom{1,}762 \\ \bottomrule
    \end{tabular}}
    \caption{Statistics of the extension of CoMMuTE.}
    \label{tab:stats-commute}
\end{table}

%CoMMuTE \citep{futeral-etal-2023-tackling} is a contrastive test set whose aim is to gauge how well MMT models use images to disambiguate ambiguous sentences and their translations. For an English sentence built around an ambiguous word, there are two translations, each with its own image, chosen to disambiguate the source sentence such that one translation is correct and the other incorrect.
Currently available for English-to-\{French,German,Czech\}, we extend the CoMMuTE benchmark to three new target languages: Arabic, Chinese and Russian, using professional translators. %\footnote{Professional translators translated the English sentence for each of the two images.} 
We also release a small validation set of 30 English ambiguous words (non-overlapping with test set examples) %\footnote{Not overlapping with those from the test set.} 
with two French translations, each with its own image, to be used for model selection during training. Table~\ref{tab:stats-commute} shows statistics of our extension of CoMMuTE.\footnote{Tokenisation using NLLB \citep{costa2022no}.}

\section{Our Approach}

Our goal is to train an MMT model capable of effectively using images to disambiguate ambiguous translations
(i.e.~where an image is necessary to translate correctly, which is MMT's main purpose)
%when necessary,
while keeping the general MT capacity of the underlying MT model, without using fully supervised data (i.e.~in a zero-shot way). This allows us to extend MMT to more language pairs, currently not possible without collecting fully supervised data. 

As shown in Figure~\ref{fig:main-fig}, we start from a strong pretrained NLLB \citep{costa2022no} MT model 
%(different versions of NLLB \citep{costa2022no} depending on the chosen model size), 
and use it to translate English captions.
%into the target languages. 
Similarly to \citet{futeral-etal-2023-tackling}, we turn it into an MMT model by adding lightweight trainable modules (visual projectors and adapters), keeping original weights frozen during training. We use visual embeddings from SigLIP \citep{zhai2023sigmoid} and concatenate them to the sequences of text embeddings in the NLLB encoder. Our approach, ZeroMMT, is based on two objectives: (1)~force the model to use images when translating by using visually-conditioned masked language modeling (VMLM) and (2)~maintain the performance of the original MT system without any fully supervised data using the Kullback-Leibler (KL) divergence between the MMT system's output and the original MT system's output distributions using the previously automatically translated data. While (1)~has already been proved successful for learning visual disambiguation capabilities \citep{futeral-etal-2023-tackling}, we further show (Section~\ref{ss:ablation}) that (2)~is key for retaining a strong translation capacity in an MMT setting, enabling zero-shot training.

In more detail, let $x_{1, \dots, n}$ denote the sequence of tokens of the English sentence, $i$ the image embedding, $y_{1, \dots, m}$ the translated sequence of tokens, $f_{\theta}$ the original MT system and $f_{\theta, \beta}$ the MMT system built on top of the text-only MT model with additional light-weight modules $\beta$, both outputting probability distributions over tokens. We formally define the losses as follows:
%\vspace{-1mm}
\resizebox{\linewidth}{!}{
\begin{minipage}{1.07\linewidth}
\begin{align}
    &\mathcal{L}_{\textit{VMLM}} = \sum_{j} y_{j} \log \big ( f_{\theta, \beta}(y_{j}; y_{<j}, x_{\setminus \mathcal{M}}, i) \big ) \\
    &\mathcal{L}_{\textit{KL}} = \sum_{j} f_{\theta}(y_j; y_{<j}, x) \log \frac{f_{\theta}(y_j; y_{<j}, x)}{f_{\theta, \beta}(y_j; y_{<j}, x, i)}
\end{align}
\end{minipage}
} \vspace{.5mm}

where $\mathcal{M}$ is the set of masked input indices. %, $\beta$ are the lightweight added trainable parameters while $\theta$ are the frozen weights of the original MT model. 
The final loss is a weighted combination of (1) and (2), and we choose the $\lambda$ value based on results on validation sets as described in Section~\ref{ss:training-details}:
%\vspace{-5mm}
\[
\mathcal{L} = \mathcal{L}_{\textit{VMLM}} + \lambda \mathcal{L}_{\textit{KL}}
\]

\section{Experiments}
\subsection{Data}
We trained our models on the Conceptual Captions dataset\footnote{At the time of writing, we were able to collect 2,831,746 out of the 3,300,000 images.} \citep{sharma-etal-2018-conceptual}. We translated Conceptual Captions into French, German, Czech, Chinese, Russian and Arabic using NLLB \citep{costa2022no} (of size 600M, 1.3B or 3.3B depending on the experiment) using a beam of size 4 for the 600M model and 2 for the largest ones.

We evaluate our models on the M30K test sets \citep{elliott-etal-2016-multi30k, elliott-EtAl:2017:WMT, barrault-etal-2018-findings} for English-to-\{German,French,Czech\}, the EMMT test set \citep{zhu-etal-2023-beyond} for English-to-Chinese, comprising 500 English product titles from e-commercial websites translated into Chinese,  and the VATEX test set \citep{wang2019vatex} for English-to-Chinese, composed of 10-second videos\footnote{We take 5 frames per second, compute SIGLIP features and average them to obtain the visual input.} with English captions translated into Chinese. We use these test sets to make sure general translation quality is not harmed when introducing additional visual inputs in unambiguous cases (as described previously, they cannot be used in practice to evaluate MMT models' ability to use images correctly). Finally we evaluate on CoMMuTE %\citep{futeral-etal-2023-tackling} 
for English-to-\{German,French,Czech,Chinese,Russian,Arabic\}, used to test how well the MMT models exploit visual context for disambiguation.

\begin{table*}[!htbp]
    \centering\small
    \resizebox{0.75\linewidth}{!}{\begin{tabular}{lcccccc} \toprule
         & Ar & Cs & De & Fr & Ru & Zh \\ \midrule
      Text-only MT baselines  & \textit{50.0} & \textit{50.0} & \textit{50.0} & \textit{50.0} & \textit{50.0} & \textit{50.0} \\ 
      NLLB-SIGLIP \textit{topline} & 82.6 & 76.0 & 83.7 & 75.0 & 75.8 & 88.1 \\ \midrule
      & \multicolumn{6}{c}{MMT -- \textit{fully supervised}} \\ \midrule
      Gated Fusion \scriptsize \textit{bilingual} & - & 51.0 \scriptsize $\pm$1.9 & 49.7 \scriptsize $\pm$0.6 & 50.0 \scriptsize $\pm$0.8 & -  & - \\
      VTLM + MMT \scriptsize \textit{bilingual} & - & 52.0 \scriptsize $\pm$0.7 & 50.2 \scriptsize $\pm$0.3 & 51.4 \scriptsize $\pm$0.9 & -  & -\\
      VGAMT \scriptsize \textit{full bilingual} & - & \underline{55.6} \scriptsize $\pm$0.8 & \textbf{59.0} \scriptsize $\pm$0.5 & \textbf{67.1} \scriptsize $\pm$0.7 & - & - \\
      VGAMT \small \textit{SIGLIP-only multi.} & - & \textbf{57.5} \scriptsize $\pm$1.2 & \underline{57.1} 
      \scriptsize $\pm$0.4 & \underline{61.3} \scriptsize $\pm$1.1 & - & - \\ \midrule
      & \multicolumn{6}{c}{MMT -- \textit{cross-lingual zero-shot}} \\ \midrule
      M2KT-VPN \scriptsize \textit{bilingual} & - & 50.1 \scriptsize $\pm$0.6 & 50.3 \scriptsize $\pm$0.5 & 50.9 \scriptsize $\pm$0.8 & - & - \\ \midrule
      & \multicolumn{6}{c}{MMT -- \textit{zero-shot}} \\ \midrule
      Multilingual OpenFlamingo & \textbf{61.3}\hphantom{\scriptsize $\pm$1.2} & 59.1\hphantom{\scriptsize $\pm$1.2} & \textbf{63.7}\hphantom{\scriptsize $\pm$1.2} & \textbf{68.5}\hphantom{\scriptsize $\pm$1.2} & \textbf{67.4}\hphantom{\scriptsize $\pm$1.2} & \textbf{66.5}\hphantom{\scriptsize $\pm$1.2} \\
      ZeroMMT-600M (\textit{ours}) \small \textit{multi.} & 56.1\scriptsize$\pm$0.8 & 55.5\scriptsize$\pm$0.5 & 55.7\scriptsize$\pm$0.3 & 58.7\scriptsize$\pm$0.4 & 57.2\scriptsize$\pm$1.2 & 58.2\scriptsize$\pm$1.1 \\
      
      ZeroMMT-1.3B (\textit{ours}) \small \textit{multi.} & 57.3\scriptsize$\pm$0.2
      & \underline{59.4}\scriptsize$\pm$0.5 & 57.4\scriptsize$\pm$0.4 & 62.2\scriptsize$\pm$0.5 & 60.6\scriptsize$\pm$0.5 & \underline{60.1}\scriptsize$\pm$0.8 \\
      
      ZeroMMT-3.3B (\textit{ours}) \small \textit{multi.} & \underline{58.9}\scriptsize$\pm$0.5
      & \textbf{61.7}\scriptsize$\pm$0.3 & \underline{60.8}\scriptsize$\pm$0.8 & \underline{65.0}\scriptsize$\pm$0.7 & \underline{62.9}\scriptsize$\pm$0.3 & \underline{60.1}\scriptsize$\pm$0.7 \\
      \bottomrule
    \end{tabular}}
    \caption{Results on CoMMuTE, averaged over 3 runs ($\pm$ standard error). The best scores for each category are in \textbf{bold} and the second best are \underline{underlined}.}
    \label{tab:commute-res}
\end{table*}

\begin{table*}[!ht]
    \centering\small
    \resizebox{0.9\linewidth}{!}{\centering\small
    \begin{tabular}{lccccccccc} \toprule
        % & \multicolumn{8}{c}{Generation} && CoMMuTE \\  \cmidrule(c){2-9} \cmidrule(c){11-11}
         & \multicolumn{2}{c}{Fr} & \multicolumn{2}{c}{De} & \multicolumn{2}{c}{Cs} & \multicolumn{2}{c}{Zh} \\ % && \multirow{2}{*}{Accuracy} \\ 
         & BLEU & COMET & BLEU & COMET & BLEU & COMET & BLEU & COMET \\ \midrule %&& \\ \midrule
         \multicolumn{9}{c}{Text-only MT baselines} \\ \midrule
        NLLB-600M \textit{distilled}& 49.17 \scriptsize $\pm$0.78 & 85.18 \scriptsize $\pm$0.67 & 33.04 \scriptsize $\pm$3.44 & 81.98 \scriptsize $\pm$2.16 & 26.58 \scriptsize $\pm$0.19 & 85.02 \scriptsize $\pm$0.42 & 16.07 \scriptsize $\pm$1.35 & 57.81 \scriptsize $\pm$4.21\hphantom{0} \\
        
        NLLB-1.3B & 51.90 \scriptsize $\pm$0.79 & 86.28 \scriptsize $\pm$0.77 & 35.39 \scriptsize $\pm$2.83 & 83.49 \scriptsize $\pm$2.15 & 30.77 \scriptsize $\pm$0.46 & 87.48 \scriptsize $\pm$0.29 & 18.22 \scriptsize $\pm$0.20 & 60.02 \scriptsize $\pm$3.51\hphantom{0} \\

        NLLB-3.3B & 53.73 \scriptsize $\pm$0.57 & 86.98 \scriptsize $\pm$0.88 & 37.26 \scriptsize $\pm$2.10 & 84.76 \scriptsize $\pm$1.76 & 33.37 \scriptsize $\pm$0.27 & 88.70 \scriptsize $\pm$0.37 & 20.55 \scriptsize $\pm$0.46 & 61.27 \scriptsize $\pm$3.50\hphantom{0} \\ \midrule % && \multirow{2}{*}{0.50} \\ 
        %\textit{distilled} & & & & & & & & \\ \midrule
        \multicolumn{9}{c}{MMT -- \textit{fully supervised}} \\ \midrule

        Gated Fusion \scriptsize \textit{bilingual} & 49.79 \scriptsize $\pm$7.46 & 80.62 \scriptsize $\pm$3.01 & 31.57 \scriptsize $\pm$5.24 & 72.89 \scriptsize $\pm$3.15 & 28.30 \scriptsize $\pm$2.52 & 79.24 \scriptsize $\pm$2.41 & - & - \\ % && 49.70 ($\pm$1.74) \\ 
        
        VTLM + MMT \scriptsize \textit{bilingual} & 55.27 \scriptsize $\pm$6.00 & 83.45 \scriptsize $\pm$1.98 & 35.94 \scriptsize $\pm$3.44 & 79.10 \scriptsize $\pm$2.35 & 32.63 \scriptsize $\pm$2.26 & 82.40 \scriptsize $\pm$1.77 & - & - \\
        
        VGAMT \scriptsize \textit{full bilingual} & 59.97 \scriptsize $\pm$6.66 & 88.29 \scriptsize $\pm$1.83 & 39.10 \scriptsize $\pm$3.14 & 85.72 \scriptsize $\pm$1.73 & 35.89 \scriptsize $\pm$1.70 & 89.50 \scriptsize $\pm$1.08 & - & - \\
        %&& 60.58 ($\pm$4.86) \\
        VGAMT \scriptsize \textit{SIGLIP-only multi.} & 58.39 \scriptsize $\pm$5.67 & 87.27 \scriptsize $\pm$1.74 & 37.36 \scriptsize $\pm$3.51 & 83.85 \scriptsize $\pm$2.04 & 34.88 \scriptsize $\pm$1.77 & 87.45 \scriptsize $\pm$1.19 & - & - \\ \midrule %&& 58.61 ($\pm$2.10) \\ \midrule

         \multicolumn{9}{c}{MMT -- \textit{cross-lingual zero-shot}} \\ \midrule

         M2KT-VPN \scriptsize \textit{bilingual} & 51.58 \scriptsize $\pm$6.72 & 80.19 \scriptsize $\pm$3.77 & 29.27 \scriptsize $\pm$5.77 & 71.63 \scriptsize $\pm$2.77 & 28.02 \scriptsize $\pm$2.31 & 78.63 \scriptsize $\pm$2.77 & - & - \\  \midrule
        
        \multicolumn{9}{c}{MMT -- \textit{zero-shot}} \\ \midrule

        Multilingual OpenFlamingo & 35.08 \scriptsize $\pm$0.76 & 82.66 \scriptsize $\pm$1.38 & 24.92 \scriptsize $\pm$2.89 & 79.93 \scriptsize $\pm$2.44 & \hphantom{0}3.27 \scriptsize $\pm$0.04 & 70.73 \scriptsize $\pm$0.55 & \hphantom{0}8.60 \scriptsize $\pm$5.86 & 53.38 \scriptsize $\pm$10.24 \\ %&& \\
        %Ours bilingual & 49.27 ($\pm$1.16) & 84.91 ($\pm$0.55)\hphantom{0} & 32.49 ($\pm$3.48) & & & & & && \\
        ZeroMMT-600M (\textit{ours}) \scriptsize \textit{multi.} & 49.00 \scriptsize $\pm$1.07 & 84.82 \scriptsize $\pm$0.79 & 32.79 \scriptsize $\pm$2.97 & 81.13 \scriptsize $\pm$2.48 & 25.24 \scriptsize $\pm$0.62 & 83.79 \scriptsize $\pm$0.55 & 15.74 \scriptsize $\pm$1.62 & 57.10 \scriptsize $\pm$4.72\hphantom{0} \\
        
        ZeroMMT-1.3B (\textit{ours}) \scriptsize \textit{multi.} & 52.06 \scriptsize $\pm$1.15 & 86.15 \scriptsize $\pm$0.84 & 35.18 \scriptsize $\pm$2.58 & 83.35 \scriptsize $\pm$1.90 & 30.14 \scriptsize $\pm$0.48 & 86.94 \scriptsize $\pm$0.33 & 17.11 \scriptsize $\pm$0.71 & 59.17 \scriptsize $\pm$4.34\hphantom{0} \\
        
        ZeroMMT-3.3B (\textit{ours}) \scriptsize \textit{multi.} & 53.34 \scriptsize $\pm$0.50 & 86.69 \scriptsize $\pm$0.94 & 37.08 \scriptsize $\pm$2.49 & 84.41 \scriptsize $\pm$1.77 & 33.03 \scriptsize $\pm$0.34 & 88.37 \scriptsize $\pm$0.32 & 19.43 \scriptsize $\pm$0.64 & 60.61 \scriptsize $\pm$4.28\hphantom{0} \\ \bottomrule %&& \\ \bottomrule
    \end{tabular}}
    \caption{Aggregated generation results for En$\rightarrow$X. Fr and De results are averaged over Test2016, Test2017 from M30K and AmbiguousCOCO. Cs results are averaged over M30K Test2016 and Test2018. Zh results are averaged over EMMT and VATEX test sets.}
    \label{tab:mmt-res}
\end{table*}

\subsection{Implementation details} \label{ss:training-details}

\paragraph{Modelling}
We trained three different versions of ZeroMMT depending on the size of the underlying NLLB model\footnote{As implemented in \texttt{Transformers} \citep{wolf-etal-2020-transformers}.} \citep{costa2022no} (600M, 1.3B and 3.3B). For SigLIP \citep{zhai2023sigmoid}, we use \texttt{ViT-B-16-SigLIP-384} trained on WebLI \citep{chen2023pali} from the \texttt{timm} library \citep{rw2019timm}. Following VGAMT \citep{futeral-etal-2023-tackling}, we used bottleneck adapters \citep{houlsby2019parameter} as implemented in the Adapters Python library \citep{poth-etal-2023-adapters} with a factor reduction of 8 and ReLU activation \citep{agarap2018deep} for each layer. The visual projector is a 1-layer neural network followed by ReLU activation projecting SigLIP \citep{zhai2023sigmoid} embeddings towards the hidden dimension of NLLB. The image representation is then concatenated to the sequence of text embeddings. The cross-attention mechanism in the decoder of the model can only attend to the positions of text embeddings. Similarly to VGAMT, we randomly mask 25\% of the input tokens for VMLM.

\paragraph{Training}
We train our models with a batch size of 32, the Adam optimizer \citep{2015-kingma} with $\beta_1 = 0.9$ and $\beta_2 = 0.99$ and learning rate of $10^{-4}$. We use $\lambda = 0.1$ to balance the two training losses. All hyperparameters were selected based on the combination of the CoMMuTE validation set (see Section~\ref{ss:commute}) and the English--French validation dataset of M30K, each score weighted equally. All our models are multilingual if not otherwise specified. We run each experiment three times with three different seeds and report average scores and standard error. It took 15 hours on one NVIDIA V100 for the 600M model and 20 hours on one NVIDIA A100 for the largest models.

\paragraph{Evaluation}
We evaluate MMT generation with BLEU \citep{papineni-etal-2002-bleu} and COMET \citep{rei-etal-2020-comet}. For BLEU, we use the Sacrebleu implementation \citep{post-2018-call} with \textit{13a} tokenization for French, German and Czech and \textit{zh} tokenization for Chinese. For COMET, we use the \texttt{wmt22-comet-da} \citep{rei-etal-2022-comet} model from the XLM-R backbone \citep{conneau-etal-2020-unsupervised}. The translations were obtained with beam search decoding of size 4. Following \citep{futeral-etal-2023-tackling}, we calculate disambiguation accuracy using CoMMuTE: given an English sentence and an associated image, we compute the perplexities of each contrastive translation, giving a score of 1 if the perplexity of the correct translation is lower than the perplexity of the contrastive one and 0 otherwise.

\subsection{Results} \label{ss:results}

\paragraph{Baselines and comparative models}
We compare our approach to several others. Firstly, we compare to the text-only MT systems on which the ZeroMMT models are based, NLLB-600M \textit{distilled}, NLLB-1.3B and NLLB-3.3B. We also compare against well-known fully supervised MMT systems: Gated Fusion \citep{wu-etal-2021-good}, a tiny 3M-parameter Transformer \citep{vaswani2017attention} trained from scratch on M30K; VTLM (+ MMT) \citep{caglayan-etal-2021-cross}, a 44M-parameter MMT system first pretrained on the translation language modelling (TLM) objective \citep{lample2019cross} with an additional image as input on translated captioning data (using the same translated data and tokenizer as ZeroMMT-600M) and then MMT-finetuned on M30K; and VGAMT \citep{futeral-etal-2023-tackling}, a 630M-parameter MMT system (of which 13M trainable), which is an MT-fine-tuned mBART \citep{liu-etal-2020-multilingual-denoising} transformed into an MMT system through the addition of lightweight adapters trained jointly on the MMT and VMLM objectives. VGAMT originally uses multiple types of visual input and is bilingual. Therefore, to have a comparable setup we retrain a VGAMT-like model with NLLB-600M \textit{distilled} as the underlying MT model, with SIGLIP features only and in a multilingual setting. Finally, we compare to Multilingual OpenFlamingo \citep{futeral2024moscar}, a 3B multilingual multimodal language model pretrained on a large number of text-image pairs and interleaved documents which allows for zero-shot MMT in a way that is comparable with our model and M2KT-VPN \citep{hirasawa-etal-2023-visual}, a cross-lingual zero-shot MMT model based on a tiny Transformer. 

We compute an approximate upperbound on CoMMuTE for models trained with SIGLIP features by evaluating on NLLB-SIGLIP \citep{visheratin2023nllb}. For each CoMMuTE instance, we compute the cosine similarity between the translation and (i)~its associated image and (ii)~the other image. If the cosine similarity of (i) is higher than (ii), it is considered a correct prediction.

\begin{table}[t]
    \centering
    \resizebox{\linewidth}{!}{\begin{tabular}{lcccccc} \toprule
                    &  Ar & Cs & De & Fr & Ru & Zh \\ \midrule
       NLLB-600M     & 79.06 & \textbf{83.31} & 80.84 & 80.17 & 79.94 & 75.41 \\
       ZeroMMT-600M (\textit{ours})          & \textbf{79.43} & 82.62 & \textbf{81.11} & \textbf{81.36} & \textbf{80.67} & \textbf{75.90} \\ \midrule
       NLLB-1.3B & 80.59 & 84.12 & 81.28 & 80.14 & 81.72 & 75.30 \\
       ZeroMMT-1.3B & \textbf{81.02} & \textbf{85.41} & \textbf{83.46} & \textbf{82.35} & \textbf{83.45} & \textbf{77.06} \\ \midrule
       NLLB-3.3B & 79.90 & 84.97 & 81.20 & 81.16 & 82.26 & 76.69 \\
       ZeroMMT-3.3B & \textbf{80.64} & \textbf{86.69} & \textbf{83.54} & \textbf{83.40} & \textbf{83.87} & \textbf{77.85} \\
       \bottomrule
    \end{tabular}}
    \caption{COMET scores on CoMMuTE (as a translation set). The best result for each model size is in \textbf{bold}.}
    \label{tab:tr-commute}
\end{table}

\paragraph{Quantitative results}
\Cref{tab:commute-res,tab:mmt-res} show the results on CoMMuTE and the aggregated results on generation benchmarks not composed of ambiguous examples (i.e.~images are not required) respectively. For full results, see Appendix~\ref{sec:detailed-results}. Compared to the text-only NLLB-600M distilled model, our approach results in only a small drop in performance on generation benchmarks (-0.52 BLEU and -0.79 COMET on average), where images are not required to translate the sentence correctly, despite not using the M30K training data or any fully supervised data. For the disambiguation task, Multilingual OpenFlamingo obtains the strongest CoMMuTE scores but it fails in generation as it was not specifically trained to translate. Our approach is significantly better than the random baseline ($>$55\% for all languages for the smallest model, $>$61\% on average for the largest model), showing that it is able to exploit images for disambiguation; results are close to VGAMT scores (for similar model sizes) for Czech despite not having been trained on fully supervised data. Table~\ref{tab:tr-commute} shows additional results on CoMMuTE but used as a traditional MMT generation benchmark. We obtain higher COMET scores than the text-only baseline NLLB on all languages for all model sizes except Czech for the smallest model. These results show that our approach is able to improve translation performance by exploiting images for disambiguation in cases of ambiguous examples without using any fully supervised data during training. We shall see in Section~\ref{sec:cfg} how image exploitation can be controlled at inference time and how our approach can be made to outperform Multilingual OpenFlamingo in image exploitation (as measured on CoMMuTe).

%\vspace{-5mm}

\begin{figure}[!hbtp]
%\resizebox{\linewidth}{*}{
\centering
\begin{subfigure}[!htbp]{0.48\textwidth}
   \centering
    \includegraphics[width=0.9\linewidth]{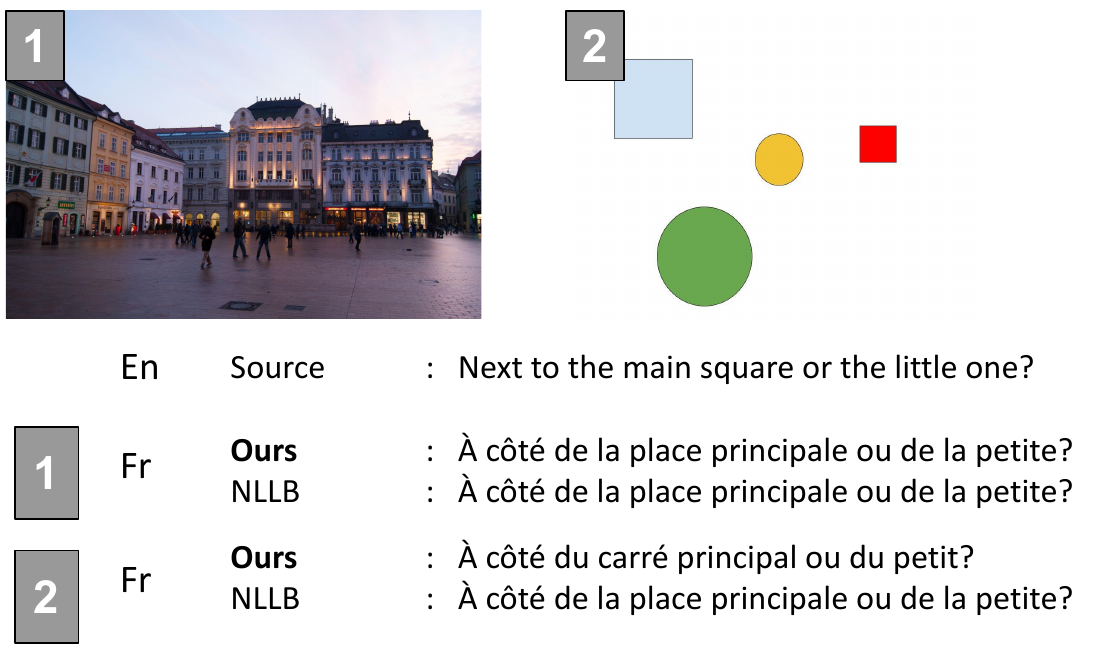}
    \caption{English--French example.}
    \label{fig:fr-ex} 
\end{subfigure}\vspace{0.05\linewidth}%

\begin{subfigure}[!t]{0.48\textwidth}
    \centering
    \includegraphics[width=.85\linewidth]{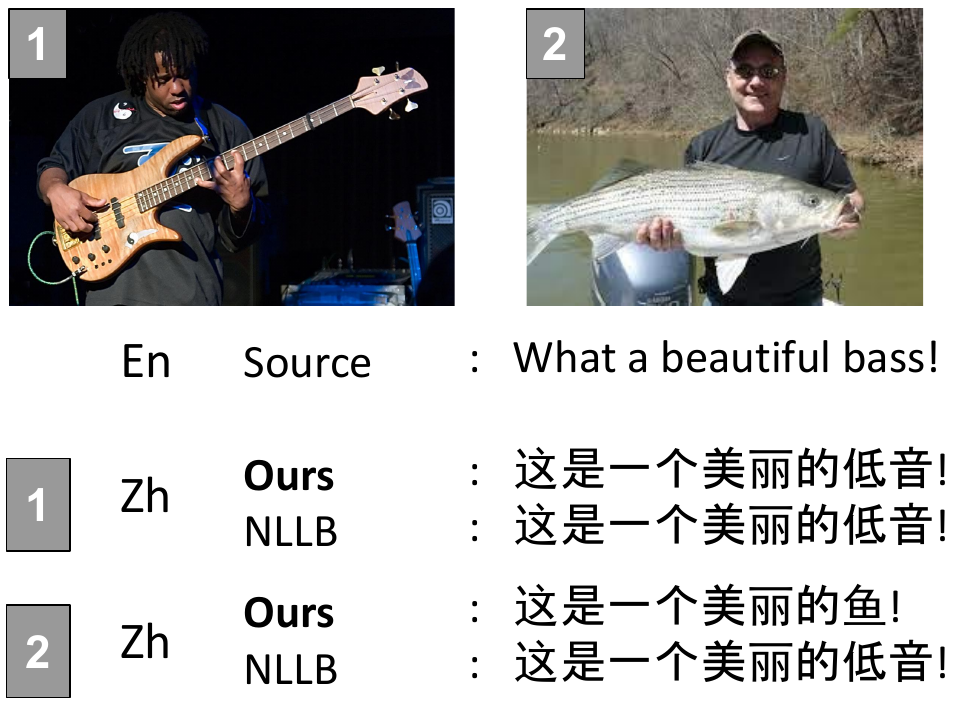}
    \caption{English--Chinese example.}
    \label{fig:zh-ex} %\vspace{0.05\linewidth}
\end{subfigure}\vspace{0.05\linewidth}%

\caption{Translations of CoMMuTE by our approach, ZeroMMT-600M, and the NLLB distilled MT model.}
\end{figure}

\paragraph{Qualitative results}

We analysed some translations of our ZeroMMT-600M model and compared them with those of the text-only distilled NLLB-600M model. %We notice that 
Our model is able to exploit the image to slightly change the translation towards the correct meaning, as shown in Figure~\ref{fig:fr-ex}, where ambiguous parts of the translations change when the image is provided. In Figure~\ref{fig:zh-ex}, the translation is also improved; \textit{bass} is translated as \zh{鱼} `fish' rather than \zh{低音} `bass (low tone)'.
%
%In the example (2) of Figure~\ref{fig:cs-ex}, the model exploits the image to change the translation from \textit{trenéry} `trainer' to \textit{autobusy} `bus', which, although not correct, is closer to the reference translation. Again, in Figure~\ref{fig:zh-ex}, the translation is improved, with the word \textit{bass} being translated as \zh{鱼} `fish' rather than as \zh{低音} `bass (low tone)'.
%
We also notice few variations in the other areas of the translation with respect to the NLLB translation, which means that our model correctly identifies the part to change. More examples can be found in Appendix~\ref{ss:additional-examples}.

\begin{figure}[t]
    \centering
    \includegraphics[width=0.8\linewidth]{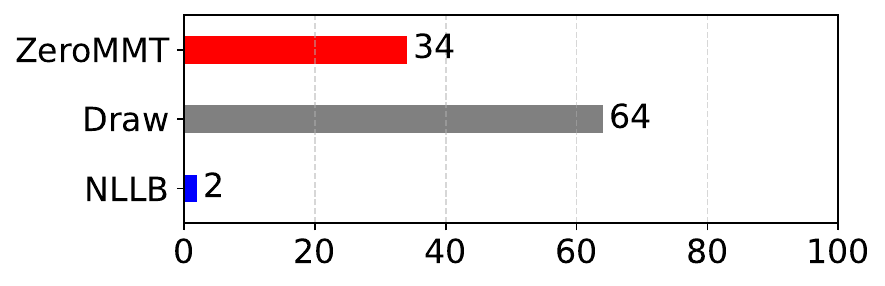}
    \caption{French human evaluation on 100 examples from CoMMuTE. `Draw' means both translations are exactly the same or considered of same quality.}
    \label{fig:human-eval-commute-fr}
\end{figure}

\begin{figure}[t]
    \centering
    \includegraphics[width=0.8\linewidth]{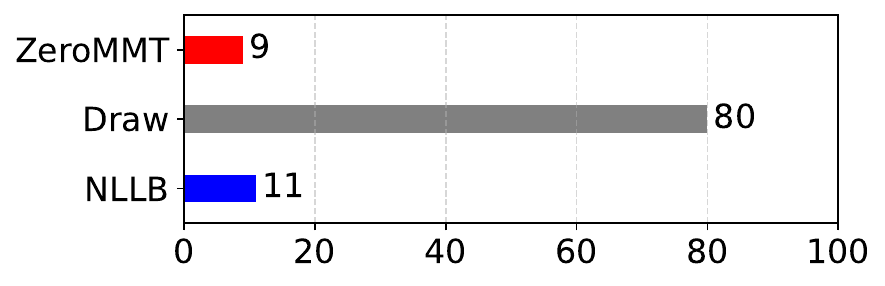}
    \caption{French human evaluation on 100 examples from M30K test sets. `Draw' means both translations are exactly the same or considered of the same quality.}
    \label{fig:human-eval-m30k-fr}
\end{figure}

\begin{figure*}[htp]
    \begin{minipage}{0.48\textwidth}
        \centering
        \includegraphics[width=.9\textwidth]{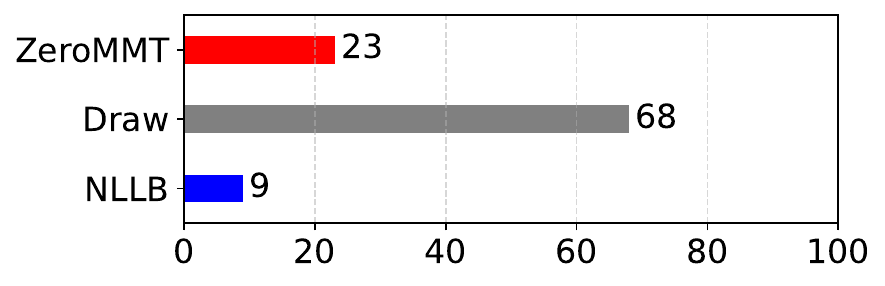}  % Replace with your image
        \caption{Chinese human evaluation on 100 examples from CoMMuTE. `Draw' means both translations are exactly the same or considered of same quality.}
        \label{fig:zh-human-eval-commute}
    \end{minipage}
    \hfill
    \begin{minipage}{0.48\textwidth}
        \centering
        \includegraphics[width=.9\textwidth]{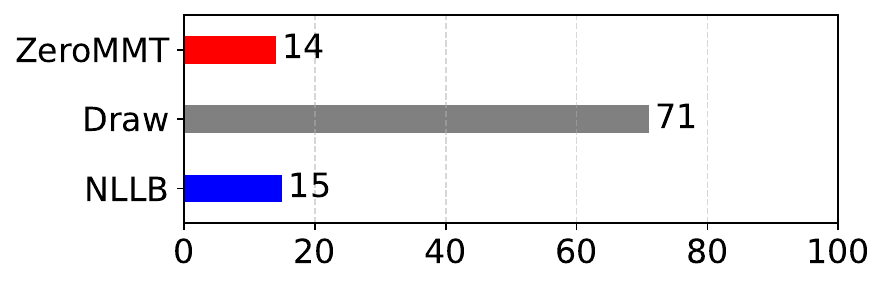}  % Replace with your image
        \caption{Chinese human evaluation on 100 examples from VATEX. `Draw' means both translations are exactly the same or considered of same quality.}
        \label{fig:zh-human-eval-vatex}
    \end{minipage}
\end{figure*}

\begin{figure}[htp]
    \centering
    \includegraphics[width=0.9\linewidth]{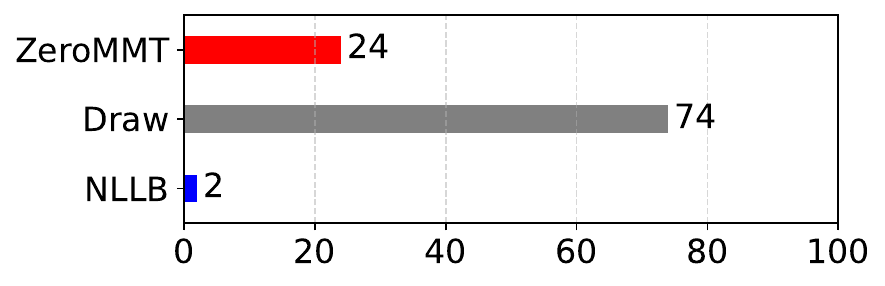}
    \caption{Arabic human evaluation on 100 examples from CoMMuTE. `Draw' means both translations are exactly the same or considered of the same quality.}
    \label{fig:ar-commute-human-eval}
\end{figure}

\paragraph{Human evaluation} We conduct human evaluation between ZeroMMT-3.3B and NLLB-3.3B outputs to further confirm these results. We set up A/B testing where annotators were asked to assess which of the translations was better given the source translation and the accompanying image. We randomly sampled 100 examples from CoMMuTE to assess which model is better in cases of ambiguity in the source sentence and 100 examples from M30K test sets (equally represented) in cases where images do not provide additional information for translation. Figure~\ref{fig:human-eval-commute-fr} shows that ZeroMMT-3.3B is considered to be better than NLLB-3.3B by a large margin in cases of ambiguity. In cases where the image adds no additional information (i.e.~the sentence to translate is non-ambiguous), Figure~\ref{fig:human-eval-m30k-fr} shows that translations are considered the same in most cases (80\%), and in the remaining cases NLLB-3.3B is considered to be slightly better than ZeroMMT-3.3B (11 vs. 9). The results of a similar analysis for Arabic and Chinese is given in \Cref{fig:zh-human-eval-commute,fig:zh-human-eval-vatex,fig:ar-commute-human-eval}. By manually looking at some examples, we noticed that the few unambiguous cases where NLLB-3.3B is considered superior to ZeroMMT-3.3B are due to hallucinations (since we add additional information in cases it is not necessary, this can occasionally occur).

\iffalse
\begin{figure*}[p]
\begin{subfigure}[!htbp]{0.45\textwidth}
    \centering
    \includegraphics[width=\linewidth]{images/example_ru.pdf}
    \caption{English--Russian example.}
    \label{fig:ru-ex}
\end{subfigure}
%
\begin{subfigure}[!htbp]{0.45\textwidth}
    \centering
    \includegraphics[width=\linewidth]{images/example_cs.pdf}
    \caption{English--Czech example.}
    \label{fig:cs-ex}
\end{subfigure}
%
\begin{subfigure}[!htbp]{0.45\textwidth}
    \centering
    \includegraphics[width=\linewidth]{images/example_zh_bass.pdf}
    \caption{English--Chinese example.}
    \label{fig:zh-ex}
\end{subfigure}
%
\begin{subfigure}[!htbp]{0.45\textwidth}
    \centering
    \includegraphics[width=\linewidth]{images/example_fr.pdf}
    \caption{English--French example.}
    \label{fig:fr-ex}
\end{subfigure}
%
\begin{subfigure}[!htbp]{0.45\textwidth}
    \centering
    \includegraphics[width=\linewidth]{images/example_arb_dough.pdf}
    \caption{English--Arabic example.}
    \label{fig:arb-ex}
\end{subfigure}
\begin{subfigure}[!htbp]{0.45\textwidth}
    \centering
    \includegraphics[width=\linewidth]{images/example_de.pdf}
    \caption{English--German example.}
    \label{fig:de-ex}
\end{subfigure}

\caption{Examples from CoMMuTE and translations by our approach and the NLLB distilled MT model.}
\end{figure*}
\fi

\section{Ablation study} \label{ss:ablation}

\begin{table}[!ht]
    \centering
    \resizebox{\linewidth}{!}{\begin{tabular}{lccc} \toprule
         & \multicolumn{2}{c}{Translation sets} & CoMMuTE\\
         & BLEU & COMET & accuracy \\ \midrule
      ZeroMMT-600M  & \underline{32.73} \scriptsize$\pm$12.33 & \underline{77.95} \scriptsize$\pm$10.82 & \underline{56.9} \scriptsize$\pm$1.4 \\ 
      \tab \textit{w/o VMLM} & \textbf{33.12} \scriptsize$\pm$12.01 & \textbf{78.49} \scriptsize$\pm$10.69 & 50.3 \scriptsize$\pm$0.4 \\ 
      \tab \textit{w/o KL} & 14.10 \scriptsize$\pm$10.70 & 65.88 \scriptsize$\pm$11.72 & \textbf{58.9} \scriptsize$\pm$1.8 \\ 
      \tab \textit{+ MMT w/o KL} & 32.09 \scriptsize$\pm$12.62 & 77.50 \scriptsize$\pm$10.80 & 55.5 \scriptsize$\pm$1.3 \\ \bottomrule
      
    \end{tabular}}
    \caption{Ablation study. Aggregated scores over benchmarks and languages. The best results are in \textbf{bold} and second best are \underline{underlined}.}
    \label{tab:ablation}
\end{table}

We conduct an ablation study on our ZeroMMT-600M model to analyse the impact of our two objectives. % on the results observed in Section~\ref{ss:results}. 
We first train a model without the VMLM objective, then a model without the KL penalty. We also test the replacement of the KL penalty with a standard auto-regressive MMT translation loss with the translated data as the ground truth, and finally we vary the KL penalty coefficient to observe the evolution of COMET and CoMMuTE scores. Additional ablation study on the choice of visual feature can be found in Appendix~\ref{ss:vis-feats}.

\paragraph{KL penalty only (i.e.~without VMLM)}
Table~\ref{tab:ablation} shows that with the KL penalty only, the model cannot exploit visual information for translation. This is because there is no need to use the input image and the model learns to ignore it. The aggregated CoMMuTE score is close to random guessing.

\paragraph{VMLM only (i.e.~without KL)}
Table~\ref{tab:ablation} also shows that, while the VMLM objective allows the model to obtain good scores on CoMMuTE (it is able to exploit visual information), the scores on generation benchmarks collapse as expected, with -19 BLEU points and -12 COMET points in comparison to the full approach.

\paragraph{KL penalty vs. MMT objective}
Finally, we replace the KL penalty with the standard MMT objective (i.e.~\textit{+MMT w/O KL} in Table~\ref{tab:ablation}) as the objective to maintain translation quality. We observe a drop of 0.64 BLEU points and 0.45 COMET points on average in comparison with the use of the KL penalty. It additionally results in an average drop of 1.4 points on CoMMuTE.

%\section{Analysis}

\paragraph{Varying the trade-off between objectives}
\begin{figure}[t]
    \centering
    \includegraphics[width=1\linewidth]{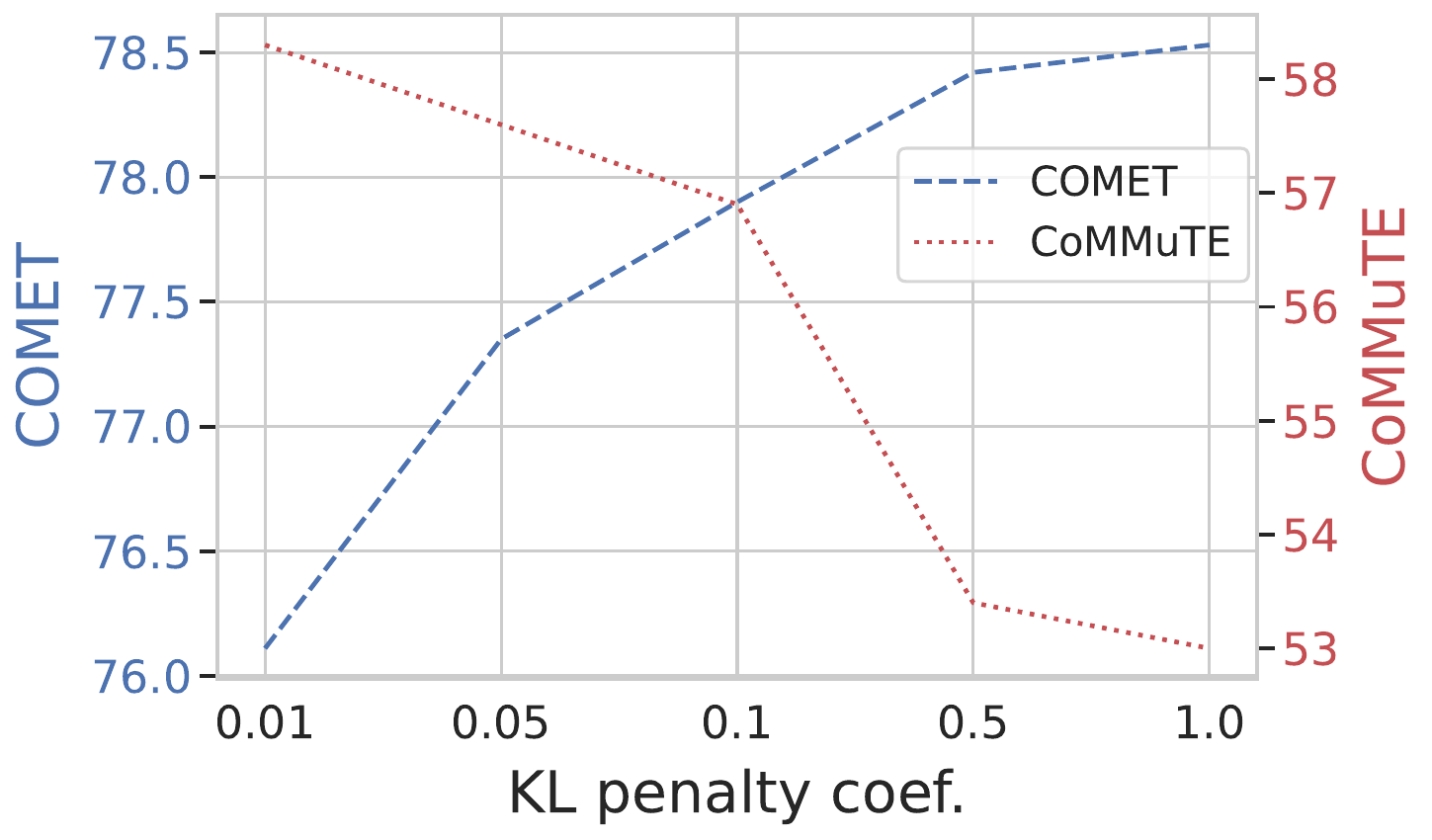}
    \caption{Evolution of aggregated COMET and CoMMuTE scores when changing the KL penalty coefficient.}
    \label{fig:kl-coef}
\end{figure}

In Figure~\ref{fig:kl-coef} we show the variation of COMET and CoMMuTE when testing our approach with different $\lambda$ coefficients for the KL penalty. We notice that when $\lambda$ is too high, it results in a large average drop of performance on CoMMuTE. 
%In contrast, the drop observed on COMET scores is relatively less important (approximately -0.5 COMET score from $\lambda$ equal to 0.1 to $\lambda$ equal to 1).

%\section{Controlling disambiguation level at inference time}
\section{Controlling the disambiguation level}
\label{sec:cfg}

\begin{table}[ht]
    \centering\small
    %\resizebox{.87\linewidth}{!}{
    \begin{tabular}{lccc} \toprule
      $\gamma$ & BLEU & COMET & CoMMuTE  \\ \midrule

         1.0 & \textbf{37.62} \scriptsize$\pm$12.11 & \textbf{81.12} \scriptsize$\pm$10.59 &  61.6 \scriptsize$\pm$2.1 \\ 
        1.25 & \underline{37.30} \scriptsize$\pm$12.07 & \underline{80.98} \scriptsize$\pm$10.56 & 64.2 \scriptsize$\pm$2.5  \\ 
        1.5 & 36.84 \scriptsize$\pm$11.89  & 80.76 \scriptsize$\pm$10.53 &  65.8 \scriptsize$\pm$2.7 \\ 
        2.0 & 35.02 \scriptsize$\pm$11.07 & 79.92 \scriptsize$\pm$10.42 & 68.5 \scriptsize$\pm$2.9  \\ 
        2.5 & 32.15 \scriptsize$\pm$10.14 & 78.42 \scriptsize$\pm$10.16 & \underline{70.3} \scriptsize$\pm$2.6  \\ 
        3.0 & 28.75 \scriptsize$\pm$9.08\hphantom{0} & 76.25 \scriptsize$\pm$9.69\hphantom{0} &  \textbf{71.7} \scriptsize$\pm$2.5 \\ \midrule
        
        {\it MOF} & {\it 20.37 \scriptsize$\pm$12.92} & {\it 73.60 \scriptsize$\pm$11.98} & {\it 64.4 \scriptsize$\pm$3.4} \\
        \bottomrule
    \end{tabular}
    %}
    \caption{Evolution of BLEU, COMET and CoMMuTE scores of our CFG-enabled ZeroMMT-3.3B model (aggregated over benchmarks and languages) compared to CFG-free ZeroMMT (i.e.~$\gamma$ = 1.0). The \textbf{best} and \underline{second best} results for each model size are shown typographically. %The second best result is \underline{underlined}. 
    ``MOF'' shows scores for Multilingual OpenFlamingo.}
    \label{tab:cfg-res}
\end{table}

We show that our method allows us to obtain an MMT system with a good trade-off between strong translation quality on unambiguous examples (i.e.~where images are not necessary to translate correctly) and the capacity to exploit visual context for disambiguation. However, some applications could require stronger disambiguation capabilities and be less reliant on translation fidelity on unambiguous cases or vice versa. Instead of retraining a model to control the trade-off between the two objectives, we instead propose to use classifier-free guidance (CFG) \citep{ho2021classifier,sanchez2023stay} to control this trade-off at inference time. We define CFG in the context of MMT as follows:
\begin{align}
\begin{split}
    \widehat{f}_{\theta, \beta}(y_j;y_{<j},x,i) = f_{\theta}(y_j;y_{<j},x) + \\
     \gamma \big( f_{\theta, \beta}(y_j;y_{<j},x,i) - f_{\theta}(y_j;y_{<j},x) \big)
\end{split}
\end{align}

\noindent where $f_{\theta}$ is the text-only MT system, $f_{\theta, \beta}$ the adapted MMT system, $x$  and $y$ the source and generated sentence, $i$ the visual input, $j$ the token index and $\gamma$ the CFG value controlling guidance.

We analyse the evolution of BLEU and COMET scores on standard generation benchmarks (where text context is enough to translate correctly), and CoMMuTE scores when varying the $\gamma$ parameter. Table~\ref{tab:cfg-res} shows that ZeroMMT-3.3B can achieve a boost in CoMMuTE accuracy of up to 4.2 points for $\gamma=$ 1.5, while facing only a moderate drop of BLEU and COMET scores on unambiguous generation benchmarks (which do not require images as additional context in theory). Higher $\gamma$ values result in stronger disambiguation capabilities, as shown by CoMMuTE, but this comes at the expense of a drop in generation quality on the unambiguous benchmarks. CFG can therefore allow us to control the trade-off between disambiguation capability and translation fidelity depending on the application. Importantly, we strongly outperform Multilingual OpenFlamingo on all metrics for different CFG values and we can obtain CoMMuTE scores up to 71.7 on average for $\gamma=$ 3.0. Results for ZeroMMT-600M and 1.3B can be found in Table~\ref{tab:cfg-res-additional} 
in Appendix~\ref{app:gamma-cfg}.

%We show in Table~\ref{tab:cfg-res} that for ZeroMMT-600M we can achieve a boost in CoMMuTE accuracy of up to 2.8 points for $\gamma=$ 1.5, while facing only a moderate drop of BLEU and COMET scores on unambiguous generation benchmarks. Higher $\gamma$ values result in stronger disambiguation capabilities, as shown by CoMMuTE, but this comes at the expense of a drop in generation quality on the unambiguous benchmarks. CFG can therefore allow us to control the trade-off between disambiguation capability and translation fidelity depending on the application. Importantly, when $\gamma \geq 2.5$, ZeroMMT-600M matches or outperforms Multilingual OpenFlamingo on both translation quality (as measured by BLEU and COMET) and image exploitation (as measured on CoMMuTE) while being five times smaller. With larger models, we strongly outperform Multilingual OpenFlamingo on all metrics for different CFG values and we can obtain CoMMuTE scores up to 71.7 on average for $\gamma=$ 3.0 for ZeroMMT-3.3B.

\section{Conclusion}
We present ZeroMMT, a novel zero-shot MMT approach bypassing the need for parallel multimodal data. ZeroMMT shows good disambiguation capabilities (it is able to effectively exploit images) while maintaining good translation results, with only a very small drop in performance according to standard generation benchmarks where images are not necessary for correct translation. ZeroMMT allows us to extend MMT to new language directions; we show that it performs well on the CoMMuTE test set for Russian and Arabic for which no parallel multimodal training data is available. Moreover, we show that it is possible to control the disambiguation-generation trade-off using classifier-free guidance. It is therefore a step towards having MMT systems that cover a broader set of languages without having to rely on acquiring costly training data.

\section*{Limitations}
While our approach allows us to exploit images for translation disambiguation as shown by the scores obtained on CoMMuTE, it is still behind the upperbound. Zero-shot disambiguation capabilities also come at the expense of a slight drop in translation quality in cases where text context is enough to translate correctly as shown by BLEU and COMET scores. To fill this gap, a next step, which we leave to future work, would be to detect ambiguity in the source sentence and access the images only in those cases. Indeed, in most cases, images are not necessary to translate the English source sentence correctly. There are therefore areas for improvement even if our zero-shot approach is close to its fully supervised counterparts. It is nevertheless a step towards zero-shot multimodal machine translation and the expansion of MMT to new language pairs.

\section*{Ethics Statement}
The released extension of CoMMuTE is designed to evaluate disambiguation capabilities of MMT systems and should not be used in any other way. Images were collected under the Creative Commons license and CoMMuTE is distributed under CC-BY-SA-4.0 license. All of our models are also distributed under CC-BY-SA-4.0 license.

\section*{Acknowledgements}
This work was granted access to the HPC resources of IDRIS under the allocation 2023-AD011013908R1 and 2023-AD011012254 made by GENCI. It was also partly funded by the last three authors’ chairs in the PRAIRIE institute funded by the French national agency ANR as part of the “Investissements d’avenir” programme under the reference ANR-19-P3IA-0001.

\bibliography{custom}
\bibliographystyle{acl_natbib}

\clearpage

\appendix

\section{Detailed results} \label{sec:detailed-results}

\subsection{Main results}
%\noindent\begin{tabular}{@{}p{0.5\textwidth-\columnsep}@{}}
\Cref{tab:m30k-full-res-fr,tab:m30k-full-res-de,tab:m30k-full-res-cs,tab:m30k-full-res-zh} show full BLEU and COMET scores for all languages and all benchmarks.
%\end{tabular}

%\FloatBarrier
\subsection{Impact of $\gamma$ in CFG}\label{app:gamma-cfg}

Table~\ref{tab:cfg-res-additional} shows the impact of $\gamma$ on BLEU, COMET and CoMMuTE scores of our CFG-enabled ZeroMMT 600M and 1.3B models (aggregated over benchmarks and languages) compared to the vanilla, CFG-free ZeroMMT model (i.e.~$\gamma$ = 1.0). See Section~\ref{sec:cfg} for results with our ZeroMMT-3.3B model.

\subsection{Ablation study - Full results}
%\noindent\begin{tabular}{@{}p{0.5\textwidth-\columnsep}@{}}
\Cref{tab:ablation-full-fr,tab:ablation-full-de,tab:ablation-full-cs,tab:ablation-full-zh} show the full results of the ablation study for all languages and all benchmarks.
%\end{tabular}

%\FloatBarrier
%\clearpage
%\twocolumn
\subsection{Ablation study - Choice of visual representation} \label{ss:vis-feats}

We additionally train several ZeroMMT-600M models with different types of visual encoder. As shown by Table~\ref{tab:vis-ft-res}, the type of visual encoder does not have an impact on global translation performances as BLEU and COMET scores do not vary a lot on standard benchmarks between models. However, we notice significant differences on CoMMuTE; the performance of CLIP,\footnote{\texttt{ViT-B-32}} SIGLIP\footnote{\texttt{vit\_base\_patch16\_siglip\_384}} and SIGLIP large\footnote{\texttt{vit\_so400m\_patch14\_siglip\_384}} visual encoders are about 1.5 points higher on average on CoMMuTE in comparison to VIT\footnote{\texttt{google/vit-base-patch16-224-in21k}} and ResNet-50.\footnote{\texttt{microsoft/resnet-50}} This is probably due to the fact that VIT and ResNet-50 are trained on ImageNet, which limits their capacity to ImageNet classes while CLIP and SIGLIP-like visual encoders are trained on free-form image-text large datasets. However, all scores on CoMMuTE are well above random, therefore validating the method for different types of visual representation.

%\onecolumn

\iffalse
\setcounter{topnumber}{10}
\setcounter{bottomnumber}{10}
\setcounter{totalnumber}{10}
\setcounter{dbltopnumber}{10}
\renewcommand\topfraction{0.99}
\renewcommand\dbltopfraction{0.99}
\renewcommand\bottomfraction{0.99}
\renewcommand\floatpagefraction{0.99}
\renewcommand\dblfloatpagefraction{0.99}
\renewcommand\textfraction{0.001}
\fi

\begin{table*}[htbp]
\centering\small
    \vspace*{1cm}
    \resizebox{0.9\linewidth}{!}{
    \begin{tabular}{lcccccc} \toprule
         & \multicolumn{2}{c}{Test2016} & \multicolumn{2}{c}{Test2017} & \multicolumn{2}{c}{COCO} \\
         & BLEU & COMET & BLEU & COMET & BLEU & COMET \\ \midrule
         
         \multicolumn{7}{c}{Text-only MT baselines} \\ \midrule
         NLLB-600M \textit{distilled} & 48.71 & 85.18 & 48.54 & 85.99 & 50.28 & 84.36 \\
         NLLB-1.3B  & 51.68 & 86.60 & 51.06 & 87.01 & 52.95 & 85.22 \\
         NLLB-3.3B  & 54.15 & 87.57 & 52.92 & 87.64 & 54.11 & 85.73 \\ \midrule

        \multicolumn{7}{c}{MMT -- \textit{fully supervised}} \\ \midrule
         Gated Fusion \textit{bilingual} & 58.70 \scriptsize$\pm$ 0.30 & 83.60 \scriptsize$\pm$ 0.08 & 50.80 \scriptsize$\pm$ 0.70 & 81.74 \scriptsize$\pm$ 0.25 & 40.40 \scriptsize$\pm$ 0.40 & 76.52 \scriptsize$\pm$ 0.25 \\
         VTLM + MMT \textit{bilingual} & 63.37 \scriptsize$\pm$ 0.13 & 85.29 \scriptsize$\pm$ 0.06 & 55.77 \scriptsize$\pm$ 0.17 & 84.35 \scriptsize$\pm$ 0.04  & 47.69 \scriptsize$\pm$ 0.16 & 80.70 \scriptsize$\pm$ 0.20 \\
         VGAMT \textit{full bilingual} & 67.20 \scriptsize$\pm$ 0.10  & 89.78 \scriptsize$\pm$ 0.04 & 61.60 \scriptsize$\pm$ 0.10  & 89.37 \scriptsize$\pm$ 0.04 & 51.10 \scriptsize$\pm$ 0.60 & 85.78 \scriptsize$\pm$ 0.11 \\
         VGAMT \textit{SIGLIP-only multi.} & 65.04 \scriptsize$\pm$ 0.52 & 88.74 \scriptsize$\pm$ 0.04 & 58.90 \scriptsize$\pm$ 0.28 & 88.23 \scriptsize$\pm$ 0.19 & 51.24 \scriptsize$\pm$ 0.73  & 84.84 \scriptsize$\pm$ 0.29 \\ \midrule

         \multicolumn{7}{c}{MMT -- \textit{cross-lingual zero-shot}} \\ \midrule

         M2KT-VPN \scriptsize \textit{bilingual} & 59.21 \scriptsize$\pm$ 0.56 & 83.95 \scriptsize$\pm$ 0.13 & 52.63 \scriptsize$\pm$ 0.63 & 81.55 \scriptsize$\pm$ 0.41 & 42.90 \scriptsize$\pm$ 0.20 & 75.08 \scriptsize$\pm$ 0.51 \\ \midrule
         \multicolumn{7}{c}{MMT -- \textit{zero-shot}} \\ \midrule
         Multilingual OpenFlamingo & 36.01 \hphantom{\scriptsize$\pm$ 0.38}& 83.56 \hphantom{\scriptsize$\pm$ 0.38}& 35.10 \hphantom{\scriptsize$\pm$ 0.38}& 83.72 \hphantom{\scriptsize$\pm$ 0.38}& 34.14 \hphantom{\scriptsize$\pm$ 0.38}& 80.71 \hphantom{\scriptsize$\pm$ 0.38}\\
         ZeroMMT-600M (\textit{ours}) \textit{multi.} & 48.62 \scriptsize$\pm$ 0.38 & 84.92 \scriptsize$\pm$ 0.09 & 48.10 \scriptsize$\pm$ 0.11 & 85.66 \scriptsize$\pm$ 0.16 & 50.29 \scriptsize$\pm$ 0.82 & 83.78 \scriptsize$\pm$ 0.20 \\
         
         ZeroMMT-1.3B (\textit{ours}) \textit{multi.} & 51.47 \scriptsize$\pm$ 0.11 & 86.42 \scriptsize$\pm$ 0.17 & 51.10 \scriptsize$\pm$ 0.02 & 87.00 \scriptsize$\pm$ 0.17 & 53.60 \scriptsize$\pm$ 0.54 & 85.03 \scriptsize$\pm$ 0.08 \\
         
         ZeroMMT-3.3B (\textit{ours}) \textit{multi.} & 52.89 \scriptsize$\pm$ 0.36 & 87.22 \scriptsize$\pm$ 0.05 & 53.29 \scriptsize$\pm$ 0.19 & 87.48 \scriptsize$\pm$ 0.13 & 53.86 \scriptsize$\pm$ 0.30 & 85.38 \scriptsize$\pm$ 0.13 \\ \bottomrule
    \end{tabular}}
    \caption{En$\rightarrow$Fr results for Test2016, Test2017 and COCO subsets of M30K, avg.~over 3 runs ($\pm$ standard error).}
    \label{tab:m30k-full-res-fr}
\end{table*}

\begin{table*}[htbp]
    \vspace*{2cm}
    \centering\small
    \resizebox{.9\linewidth}{!}{\begin{tabular}{lcccccc} \toprule
         & \multicolumn{2}{c}{Test2016} & \multicolumn{2}{c}{Test2017} & \multicolumn{2}{c}{COCO} \\
         & BLEU & COMET & BLEU & COMET & BLEU & COMET \\ \midrule
         \multicolumn{7}{c}{Text-only MT baselines} \\ \midrule
         NLLB-600M \textit{distilled} & 37.14 & 83.79 & 33.24 & 83.21 & 28.73 & 78.95 \\
         NLLB-1.3B & 37.91 & 85.14 & 36.81 & 84.86 & 31.44 & 80.45 \\
         NLLB-3.3B & 39.47 & 86.22 & 37.86 & 85.76 & 34.44 & 82.28 \\ \midrule
         
        \multicolumn{7}{c}{MMT -- \textit{fully supervised}} \\ \midrule
         Gated Fusion \textit{bilingual} & 38.70 \scriptsize$\pm$ 0.20 & 76.32 \scriptsize$\pm$ 0.17 & 29.50 \scriptsize$\pm$ 0.20 & 73.61 \scriptsize$\pm$ 0.32 & 26.60 \scriptsize$\pm$ 0.30 & 68.74 \scriptsize$\pm$ 0.36 \\
         VTLM + MMT \textit{bilingual} & 40.46 \scriptsize$\pm$ 0.64 & 81.58 \scriptsize$\pm$ 0.08 & 35.19 \scriptsize$\pm$ 0.16 & 79.79 \scriptsize$\pm$ 0.06 & 32.18 \scriptsize$\pm$ 0.21 & 75.94 \scriptsize$\pm$ 0.08 \\
         VGAMT \textit{full bilingual} & 43.30 \scriptsize$\pm$ 0.20 & 87.34 \scriptsize$\pm$ 0.08 & 38.30 \scriptsize$\pm$ 0.20 & 86.49 \scriptsize$\pm$ 0.07 & 35.70 \scriptsize$\pm$ 0.30 & 83.33 \scriptsize$\pm$ 0.08 \\
         
         VGAMT \textit{SIGLIP-only multi.} & 41.93 \scriptsize$\pm$ 0.75 & 85.79 \scriptsize$\pm$ 0.13 & 36.68 \scriptsize$\pm$ 0.23 & 84.72 \scriptsize$\pm$ 0.27 & 33.48 \scriptsize$\pm$ 0.13 & 81.05 \scriptsize$\pm$ 0.29 \\ \midrule

         \multicolumn{7}{c}{MMT -- \textit{cross-lingual zero-shot}} \\ \midrule

         M2KT-VPN \scriptsize \textit{bilingual} & 37.14 \scriptsize$\pm$ 0.69 & 75.51 \scriptsize$\pm$ 0.63 & 27.09 \scriptsize$\pm$ 0.40 & 72.29 \scriptsize$\pm$ 0.67 & 23.57 \scriptsize$\pm$ 0.54 & 66.07 \scriptsize$\pm$ 1.11 \\ \midrule
         
         \multicolumn{7}{c}{MMT -- \textit{zero-shot}} \\ \midrule

         Multilingual OpenFlamingo & 28.86 \hphantom{\scriptsize$\pm$ 0.38}& 82.31 \hphantom{\scriptsize$\pm$ 0.38} & 23.91 \hphantom{\scriptsize$\pm$ 0.38} & 80.91 \hphantom{\scriptsize$\pm$ 0.38} & 21.99 \hphantom{\scriptsize$\pm$ 0.38} & 76.58 \hphantom{\scriptsize$\pm$ 0.38} \\
         ZeroMMT-600M (\textit{ours}) \textit{multi.} & 36.22 \scriptsize$\pm$ 0.40 & 83.04 \scriptsize$\pm$ 0.39 & 33.11 \scriptsize$\pm$ 0.68 & 82.54 \scriptsize$\pm$ 0.17 & 29.04 \scriptsize$\pm$ 0.13 & 77.72 \scriptsize$\pm$ 0.16 \\
         
         ZeroMMT-1.3B (\textit{ours}) \textit{multi.} & 37.63 \scriptsize$\pm$ 0.13 & 84.80 \scriptsize$\pm$ 0.19 & 36.24 \scriptsize$\pm$ 0.54 & 84.56 \scriptsize$\pm$ 0.19 & 31.66 \scriptsize$\pm$ 0.47 & 80.68 \scriptsize$\pm$ 0.14 \\
         
         ZeroMMT-3.3B (\textit{ours}) \textit{multi.} & 39.58 \scriptsize$\pm$ 0.30 & 85.85 \scriptsize$\pm$ 0.05 & 37.97 \scriptsize$\pm$ 0.21 & 85.46 \scriptsize$\pm$ 0.13 & 33.71 \scriptsize$\pm$ 0.40 & 81.92 \scriptsize$\pm$ 0.16 \\ \bottomrule
    \end{tabular}}
    \caption{En$\rightarrow$De results for Test2016, Test2017 and COCO subsets of M30K, avg.~over 3 runs ($\pm$ standard error).}
    \label{tab:m30k-full-res-de}
\end{table*}

%~\\[1cm]

%\FloatBarrier

\begin{table*}[htbp]
    \centering\small
    \resizebox{.8\linewidth}{!}{\begin{tabular}{lcccc} \toprule
         & \multicolumn{2}{c}{Test2016} & \multicolumn{2}{c}{Test2018}  \\
         & BLEU & COMET & BLEU & COMET  \\ \midrule
         \multicolumn{5}{c}{Text-only MT baselines} \\ \midrule
         NLLB-600M \textit{distilled} & 26.39 & 85.44 & 26.76 & 84.60 \\ 
         NLLB-1.3B  & 30.31 & 87.77 & 31.23 & 87.19 \\ 
         NLLB-3.3B  & 33.64 & 89.08 & 33.10 & 88.33 \\ \midrule
        \multicolumn{5}{c}{MMT -- \textit{fully supervised}} \\ \midrule
         Gated Fusion \textit{bilingual} & 30.80 \scriptsize$\pm$ 0.40 & 81.64 \scriptsize$\pm$ 0.32 & 25.80 \scriptsize$\pm$ 0.10 & 76.85 \scriptsize$\pm$ 0.18 \\
         VTLM + MMT \textit{bilingual} & 34.87 \scriptsize$\pm$ 0.19 & 84.15 \scriptsize$\pm$ 0.17 & 30.38 \scriptsize$\pm$ 0.35 & 80.64 \scriptsize$\pm$ 0.20 \\
         VGAMT \textit{full bilingual} & 37.60 \scriptsize$\pm$ 0.20 & 90.57 \scriptsize$\pm$ 0.08 & 34.20 \scriptsize$\pm$ 0.10 & 88.43 \scriptsize$\pm$ 0.06 \\
         VGAMT \textit{SIGLIP-only multi.} & 36.62 \scriptsize$\pm$ 0.42 & 88.63 \scriptsize$\pm$ 0.16 & 33.13 \scriptsize$\pm$ 0.23 & 86.28 \scriptsize$\pm$ 0.11 \\ \midrule

         \multicolumn{5}{c}{MMT -- \textit{cross-lingual zero-shot}} \\ \midrule

         M2KT-VPN \scriptsize \textit{bilingual} & 30.29 \scriptsize$\pm$ 0.54 & 81.33 \scriptsize$\pm$ 0.50 & 25.75 \scriptsize$\pm$ 0.28 & 75.93 \scriptsize$\pm$ 0.72 \\ \midrule
         
         \multicolumn{5}{c}{MMT -- \textit{zero-shot}} \\ \midrule
         Multilingual OpenFlamingo & \hphantom{0}3.22 \hphantom{\scriptsize$\pm$ 0.38} & 71.27 \hphantom{\scriptsize$\pm$ 0.38} & \hphantom{0}3.31 \hphantom{\scriptsize$\pm$ 0.38} & 70.18 \hphantom{\scriptsize$\pm$ 0.38} \\
         ZeroMMT-600M (\textit{ours}) \textit{multi.} & 25.66 \scriptsize$\pm$ 0.43 & 84.27 \scriptsize$\pm$ 0.36 & 24.82 \scriptsize$\pm$ 0.49 & 83.32 \scriptsize$\pm$ 0.14 \\
         ZeroMMT-1.3B (\textit{ours}) \textit{multi.} & 29.98 \scriptsize$\pm$ 0.59 & 87.13 \scriptsize$\pm$ 0.27 & 30.29 \scriptsize$\pm$ 0.25 & 86.75 \scriptsize$\pm$ 0.28 \\
         ZeroMMT-3.3B (\textit{ours}) \textit{multi.} & 32.99 \scriptsize$\pm$ 0.38 & 88.67 \scriptsize$\pm$ 0.07 & 33.08 \scriptsize$\pm$ 0.30 & 88.06 \scriptsize$\pm$ 0.11 \\ \bottomrule
    \end{tabular}}
    \caption{En$\rightarrow$Cs results for Test2016 and Test2018 subsets of M30K, avg.~over 3 runs ($\pm$ standard error).}
    \label{tab:m30k-full-res-cs}
\end{table*}

\begin{table*}[htbp]
    \vspace*{1cm}
    \centering\small
    \resizebox{.8\linewidth}{!}{\begin{tabular}{lcccc} \toprule
         & \multicolumn{2}{c}{EMMT} & \multicolumn{2}{c}{VATEX}  \\
         & BLEU & COMET & BLEU & COMET  \\ \midrule
         \multicolumn{5}{c}{Text-only MT baselines} \\ \midrule
         NLLB-600M \textit{distilled} & 14.72 & 53.60 & 17.42 & 62.03 \\
         NLLB-1.3B & 18.42 & 56.51 & 18.02 & 63.54 \\ 
         NLLB-3.3B & 21.01 & 57.77 & 20.09 & 64.77 \\ \midrule
         \multicolumn{5}{c}{MMT -- \textit{zero-shot}} \\ \midrule
         Multilingual OpenFlamingo & \hphantom{0}2.74 \hphantom{\scriptsize$\pm$ 0.38} & 43.14 \hphantom{\scriptsize$\pm$ 0.38} & 14.46 \hphantom{\scriptsize$\pm$ 0.38} & 63.62 \hphantom{\scriptsize$\pm$ 0.38} \\
         ZeroMMT-600M (\textit{ours}) \textit{multi.} & 14.12 \scriptsize$\pm$ 0.09 & 52.39 \scriptsize$\pm$ 0.07 & 17.36 \scriptsize$\pm$ 0.13 & 61.82 \scriptsize$\pm$ 0.12 \\
         ZeroMMT-1.3B (\textit{ours}) \textit{multi.} & 16.42 \scriptsize$\pm$ 0.15 & 54.84 \scriptsize$\pm$ 0.44 & 17.80 \scriptsize$\pm$ 0.16 & 63.50 \scriptsize$\pm$ 0.16 \\
         ZeroMMT-3.3B (\textit{ours}) \textit{multi.} & 18.97 \scriptsize$\pm$ 0.59 & 56.34 \scriptsize$\pm$ 0.50 & 19.88 \scriptsize$\pm$ 0.25 & 64.87 \scriptsize$\pm$ 0.07 \\ \bottomrule
    \end{tabular}}
    \caption{En$\rightarrow$Zh results for EMMT and VATEX test sets, averaged over 3 runs ($\pm$ standard error).}
    \label{tab:m30k-full-res-zh}
\end{table*}

%~\\[2cm]

%\FloatBarrier

\begin{table*}[htbp]
    \vspace*{1cm}
    \centering\small
    \resizebox{.8\linewidth}{!}{\begin{tabular}{lcccccc} \toprule
      $\gamma$ & BLEU & COMET & CoMMuTE &  BLEU & COMET & CoMMuTE  \\ \midrule
      & \multicolumn{3}{c}{ZeroMMT-600M} & \multicolumn{3}{c}{ZeroMMT-1.3B} \\ \midrule
      1.0 & \textbf{32.73} \scriptsize$\pm$12.33 & \textbf{77.95} \scriptsize$\pm$10.82 &  56.9 \scriptsize$\pm$1.4 & \textbf{35.62} \scriptsize$\pm$12.58 & \textbf{80.07} \scriptsize$\pm$10.78 &  59.5 \scriptsize$\pm$1.8 \\ 
        1.25 & \underline{32.39} \scriptsize$\pm$12.24 & \underline{77.73} \scriptsize$\pm$10.76 & 58.4 \scriptsize$\pm$1.4 & \underline{35.25} \scriptsize$\pm$12.56 & \underline{79.89} \scriptsize$\pm$10.73 & 61.6 \scriptsize$\pm$1.3  \\ 
        1.5 & 31.81 \scriptsize$\pm$12.04  & 77.39 \scriptsize$\pm$10.66 &  59.7 \scriptsize$\pm$1.8 & 34.67 \scriptsize$\pm$12.42  & 79.62 \scriptsize$\pm$10.65 &  63.8 \scriptsize$\pm$2.8 \\ 
        2.0 & 30.29 \scriptsize$\pm$11.52 & 76.35 \scriptsize$\pm$10.46 & 62.3 \scriptsize$\pm$1.9 & 32.96 \scriptsize$\pm$11.92 & 78.72 \scriptsize$\pm$10.40 & 66.1 \scriptsize$\pm$2.4  \\ 
        2.5 & 27.98 \scriptsize$\pm$10.75 & 74.68 \scriptsize$\pm$10.09 & \underline{64.1} \scriptsize$\pm$2.1 & 30.36 \scriptsize$\pm$11.29 & 77.09 \scriptsize$\pm$10.04 & \underline{68.0} \scriptsize$\pm$2.4  \\ 
        3.0 & 25.03 \scriptsize$\pm$9.56\hphantom{0} & 72.29 \scriptsize$\pm$9.58\hphantom{0} &  \textbf{65.4} \scriptsize$\pm$2.2 & 27.06 \scriptsize$\pm$10.21 & 74.60 \scriptsize$\pm$9.49\hphantom{0} &  \textbf{69.2} \scriptsize$\pm$2.0 \\
        \bottomrule
    \end{tabular}}
    \caption{Evolution of BLEU, COMET and CoMMuTE scores of our CFG-enabled ZeroMMT 600M and 1.3B models (aggregated over benchmarks and languages) compared to the vanilla, CFG-free ZeroMMT model (i.e.~$\gamma=1.0$). The best result of each model size is in \textbf{bold}. The second best result is \underline{underlined}. See Section~\ref{sec:cfg} for ZeroMMT-3.3B results.}
    \label{tab:cfg-res-additional}
\end{table*}

\begin{table*}[htbp]
    \centering\small
    \resizebox{\linewidth}{!}{\begin{tabular}{lccccccc} \toprule
         & \multicolumn{2}{c}{Test2016} & \multicolumn{2}{c}{Test2017} & \multicolumn{2}{c}{COCO} & CoMMuTE \\
         & BLEU & COMET & BLEU & COMET & BLEU & COMET & Accuracy \\ \midrule

    ZeroMMT-600M & 48.62 \scriptsize$\pm$ 0.38 & \underline{84.92} \scriptsize$\pm$ 0.09 & \textbf{48.10} \scriptsize$\pm$ 0.11 & \textbf{85.66} \scriptsize$\pm$ 0.16 & \textbf{50.29} \scriptsize$\pm$ 0.82 & \underline{83.78} \scriptsize$\pm$ 0.20 & \underline{58.7} \scriptsize$\pm$0.4 \\
    
    \tab \textit{w/o VMLM} & \textbf{49.01} \scriptsize$\pm$ 0.16 & \textbf{85.09} \scriptsize$\pm$ 0.04 & \underline{47.64} \scriptsize$\pm$ 0.19 & \underline{85.59} \scriptsize$\pm$ 0.04 & \underline{49.92} \scriptsize$\pm$ 0.28 & \textbf{83.91} \scriptsize$\pm$ 0.02 & 50.0  \scriptsize$\pm$ 0.3\\ % \scriptsize$\pm$ 0.26
    
    \tab \textit{w/o KL} & 28.73 \scriptsize$\pm$ 5.97 & 78.51 \scriptsize$\pm$ 2.96 & 23.63 \scriptsize$\pm$ 5.99 & 76.95 \scriptsize$\pm$ 3.40 & 30.78 \scriptsize$\pm$ 6.48 & 76.50 \scriptsize$\pm$ 2.98 & \textbf{60.4} \scriptsize$\pm$ 1.4 \\ % 60.39 \scriptsize$\pm$ 1.38
    
    \tab \textit{+ MMT w/o KL} & \underline{48.68} \scriptsize$\pm$ 0.22 & 84.64 \scriptsize$\pm$ 0.21 & \underline{47.64} \scriptsize$\pm$ 0.35 & 85.37 \scriptsize$\pm$ 0.05 & 49.40 \scriptsize$\pm$ 0.19 & 83.11 \scriptsize$\pm$ 0.18 & 56.8 \scriptsize$\pm$ 1.7 \\ \bottomrule % 56.82 \scriptsize$\pm$ 1.74

    \end{tabular}}
    \caption{Ablation study En$\rightarrow$Fr. The best result is in \textbf{bold} and the second best result is \underline{underlined}.}
    \label{tab:ablation-full-fr}
\end{table*}

%\FloatBarrier

\begin{table*}[htbp]
    \vspace*{1cm}
    \centering\small
    \resizebox{\linewidth}{!}{\begin{tabular}{lccccccc} \toprule
         & \multicolumn{2}{c}{Test2016} & \multicolumn{2}{c}{Test2017} & \multicolumn{2}{c}{COCO} & CoMMuTE \\
         & BLEU & COMET & BLEU & COMET & BLEU & COMET & Accuracy \\ \midrule

    ZeroMMT-600M & \underline{36.22} \scriptsize$\pm$ 0.40 & \underline{83.04} \scriptsize$\pm$ 0.39 & \underline{33.11} \scriptsize$\pm$ 0.68 & \underline{82.54} \scriptsize$\pm$ 0.17 & \textbf{29.04} \scriptsize$\pm$ 0.13 & \underline{77.72} \scriptsize$\pm$ 0.16 & \underline{55.7} \scriptsize$\pm$ 0.3 \\
    
    \tab \textit{w/o VMLM} & \textbf{37.17} \scriptsize$\pm$ 0.16 & \textbf{83.59} \scriptsize$\pm$ 0.08 & \textbf{33.72} \scriptsize$\pm$ 0.21 & \textbf{83.10} \scriptsize$\pm$ 0.09 & \underline{28.14} \scriptsize$\pm$ 0.38 & \textbf{78.53} \scriptsize$\pm$ 0.21 & 50.0 \scriptsize$\pm$ 0.0 \\
    
    \tab \textit{w/o KL} & 12.42 \scriptsize$\pm$ 5.76 & 67.10 \scriptsize$\pm$ 5.04 & \hphantom{0}7.92 \scriptsize$\pm$ 4.36 & 64.92 \scriptsize$\pm$ 4.64 & \hphantom{0}8.39 \scriptsize$\pm$ 4.17 & 61.32 \scriptsize$\pm$ 4.28 & \textbf{56.8} \scriptsize$\pm$ 1.1 \\ %56.78 \scriptsize$\pm$ 1.13
    
    \tab \textit{+ MMT w/o KL} & 35.95 \scriptsize$\pm$ 0.51 & 82.58 \scriptsize$\pm$ 0.10 & 32.72 \scriptsize$\pm$ 0.31 & 82.02 \scriptsize$\pm$ 0.11 & 27.40 \scriptsize$\pm$ 0.33 & 77.13 \scriptsize$\pm$ 0.02 & 54.6 \scriptsize$\pm$ 0.6  \\ \bottomrule % 54.56 \scriptsize$\pm$ 0.62

    \end{tabular}}
    \caption{Ablation study En$\rightarrow$De. The best result is in \textbf{bold} and the second best result is \underline{underlined}.}
    \label{tab:ablation-full-de}
\end{table*}

%\FloatBarrier

\begin{table*}[htbp]
    \vspace*{1cm}
    \centering\small
    \begin{tabular}{lccccc} \toprule
         & \multicolumn{2}{c}{Test2016} & \multicolumn{2}{c}{Test2018} & CoMMuTE \\
         & BLEU & COMET & BLEU & COMET & Accuracy \\ \midrule

    ZeroMMT-600M & \underline{25.66} \scriptsize$\pm$ 0.43 & \underline{84.27} \scriptsize$\pm$ 0.36 & 24.82 \scriptsize$\pm$ 0.49 & \underline{83.32} \scriptsize$\pm$ 0.14 & \underline{55.5} \scriptsize$\pm$ 0.5 \\
    \tab \textit{w/o VMLM} & \textbf{26.49} \scriptsize$\pm$ 0.20 & \textbf{85.17} \scriptsize$\pm$ 0.07 & \textbf{26.55} \scriptsize$\pm$ 0.03 & \textbf{84.34} \scriptsize$\pm$ 0.07 & 50.1 \scriptsize$\pm$ 0.2 \\ % 50.11 \scriptsize$\pm$ 0.15
    
    \tab \textit{w/o KL} & 10.90 \scriptsize$\pm$ 5.49 & 71.97 \scriptsize$\pm$ 5.42 & \hphantom{0}8.15 \scriptsize$\pm$ 4.14 & 67.27 \scriptsize$\pm$ 5.68 & \textbf{59.1}  \scriptsize$\pm$ 0.8 \\ % 59.09  \scriptsize$\pm$ 0.79
    
    \tab \textit{+ MMT w/o KL} & 25.10 \scriptsize$\pm$ 0.27 & 83.78 \scriptsize$\pm$ 0.20 & \underline{25.43} \scriptsize$\pm$ 0.10 & 82.66 \scriptsize$\pm$ 0.10 & 54.8 \scriptsize$\pm$ 0.9 \\ \bottomrule % 54.76 \scriptsize$\pm$ 0.85

    \end{tabular}
    \caption{Ablation study En$\rightarrow$Cs. The best result is in \textbf{bold} and the second best result is \underline{underlined}.}
    \label{tab:ablation-full-cs}
\end{table*}

%\FloatBarrier

\begin{table*}[htbp]
    \vspace*{1cm}
    \centering\small
    \begin{tabular}{lccccc} \toprule
         & \multicolumn{2}{c}{EMMT} & \multicolumn{2}{c}{VATEX} & CoMMuTE \\
         & BLEU & COMET & BLEU & COMET & Accuracy \\ \midrule

    ZeroMMT-600M & \underline{14.12} \scriptsize$\pm$ 0.09 & \underline{52.39} \scriptsize$\pm$ 0.07 & \underline{17.36} \scriptsize$\pm$ 0.13 & 61.82 \scriptsize$\pm$ 0.12 & \underline{58.2} \scriptsize$\pm$ 1.1 \\
    \tab \textit{w/o VMLM} & \textbf{15.19} \scriptsize$\pm$ 0.27 & \textbf{53.60} \scriptsize$\pm$ 0.11 & \textbf{17.40} \scriptsize$\pm$ 0.10 & \underline{61.95} \scriptsize$\pm$ 0.12 & 50.1 \scriptsize$\pm$ 0.2 \\ % 50.11 \scriptsize$\pm$ 0.15
    
    \tab \textit{w/o KL} & \hphantom{0}1.12 \scriptsize$\pm$ 4.86 & 43.30 \scriptsize$\pm$ 1.06 & \hphantom{0}8.99 \scriptsize$\pm$ 4.27 & 51.01 \scriptsize$\pm$ 5.38 & \textbf{60.7}  \scriptsize$\pm$ 0.5 \\ % 60.65  \scriptsize$\pm$ 0.53
    \tab \textit{+ MMT w/o KL} & 11.89 \scriptsize$\pm$ 0.44 & 51.56 \scriptsize$\pm$ 0.63 & 16.70 \scriptsize$\pm$ 0.18 & \textbf{62.16} \scriptsize$\pm$ 0.24 & 56.5 \scriptsize$\pm$ 0.5 \\ \bottomrule % 56.45 \scriptsize$\pm$ 0.46

    \end{tabular}
    \caption{Ablation study En$\rightarrow$Zh. The best result is in \textbf{bold} and the second best result is \underline{underlined}.}
    \label{tab:ablation-full-zh}
\end{table*}

\begin{table*}[htpb]
    \vspace*{1cm}
    \resizebox{\linewidth}{!}{
    \centering\small
    \begin{tabular}{lccccccc} \toprule
        
         & \multicolumn{2}{c}{Fr} & \multicolumn{2}{c}{De} & \multicolumn{2}{c}{Cs} & \multirow{2}{*}{CoMMuTE} \\ 
         & BLEU & COMET & BLEU & COMET & BLEU & COMET & \\ \midrule %&& \\ \midrule

        ZeroMMT-600M (\textit{SIGLIP}) & \textbf{49.00} \scriptsize $\pm$1.07 & \underline{84.82} \scriptsize $\pm$0.79 & \underline{32.79} \scriptsize $\pm$2.97 & 81.13 \scriptsize $\pm$2.48 & 25.24 \scriptsize $\pm$0.62 & \underline{83.79} \scriptsize $\pm$0.55 & \underline{56.90} \scriptsize $\pm$1.40 \\
        
        ZeroMMT-600M (\textit{SIGLIP large}) & 48.77 \scriptsize $\pm$1.16 & 84.61 \scriptsize $\pm$0.77 & 32.39 \scriptsize $\pm$3.20 & 81.13 \scriptsize $\pm$2.40 & 25.35 \scriptsize $\pm$0.26 & \textbf{83.89} \scriptsize $\pm$0.40 & 56.73 \scriptsize $\pm$1.65 \\
        
        ZeroMMT-600M (\textit{CLIP}) \scriptsize  & 48.94 \scriptsize $\pm$1.08 & 84.77 \scriptsize $\pm$0.74 & 32.55 \scriptsize $\pm$3.31 & 81.11 \scriptsize $\pm$2.43 & \underline{25.44} \scriptsize $\pm$0.40 & 83.62 \scriptsize $\pm$0.65 & \textbf{57.25} \scriptsize $\pm$1.70  \\ 
        
        ZeroMMT-600M (\textit{VIT}) \scriptsize  & 48.86 \scriptsize $\pm$1.45 & 84.64 \scriptsize $\pm$0.67 & 32.62 \scriptsize $\pm$3.03 & \underline{81.17} \scriptsize $\pm$2.23 & 25.10 \scriptsize $\pm$0.37 & 83.64 \scriptsize $\pm$0.53 & 55.59 \scriptsize $\pm$0.96  \\ 
        
        ZeroMMT-600M (\textit{ResNet}) & \underline{48.98} \scriptsize $\pm$1.12 & \textbf{84.90} \scriptsize $\pm$0.71 & \textbf{32.90} \scriptsize $\pm$3.20 & \textbf{81.34} \scriptsize $\pm$2.41 & \textbf{25.56} \scriptsize $\pm$0.40 & 83.75 \scriptsize $\pm$0.42 & 55.54 \scriptsize $\pm$0.91  \\ 
        \bottomrule %&& \\ \bottomrule
    \end{tabular}}
    \caption{Impact of visual features. Aggregated generation results for En$\rightarrow$X. Fr and De results are averaged over Test2016, Test2017 from M30K and AmbiguousCOCO. Cs results are averaged over M30K Test2016 and Test2018. CoMMuTE results are averaged over languages. \textbf{Bold} is best result. \underline{Underline} is second best.}
    \label{tab:vis-ft-res}
\end{table*}

%\FloatBarrier
%\clearpage

\section{Additional examples} \label{ss:additional-examples}
\noindent\begin{tabular}{@{}p{0.5\textwidth-\columnsep}@{}}
\Cref{fig:ex-fr-sheets,fig:ex-fr-bill,fig:ex-de-gum,fig:ex-ru-match,fig:ex-ar-spirits,fig:ex-zh-jam} show additional translation examples from CoMMuTE by ZeroMMT (Ours) and the text-only NLLB-600M distilled model.
\end{tabular}

%\FloatBarrier

\begin{figure*}[!ht]
    \centering\small
\begin{subfigure}[b]{0.485\textwidth}
    \includegraphics[width=\linewidth]{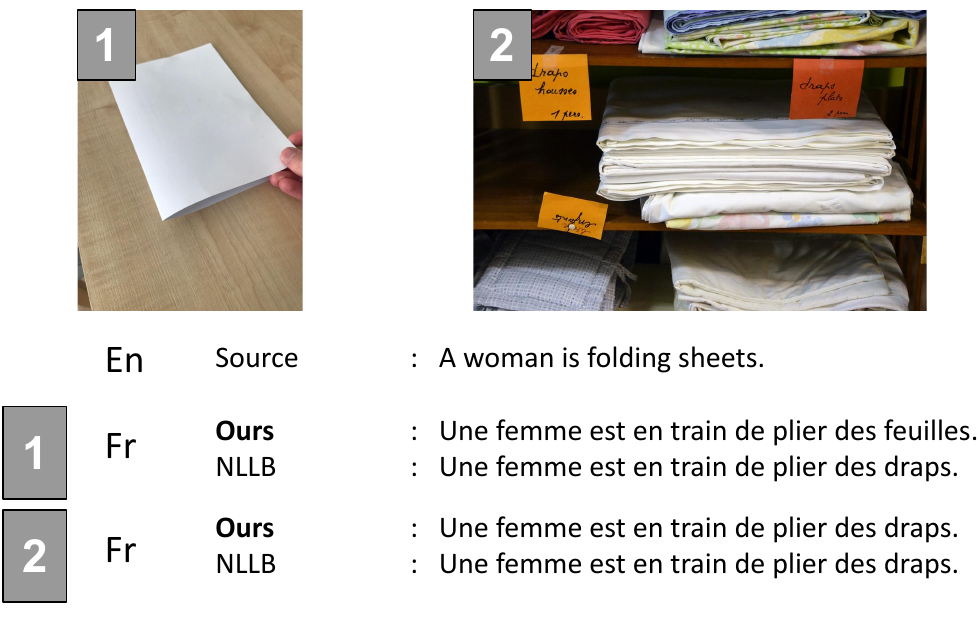}
    \caption{French example from CoMMuTE. The English word `sheets' can refer to `paper' or `bed sheets'. Correctly translated by ZeroMMT in both cases.}
    \label{fig:ex-fr-sheets}
\end{subfigure}
\hfill
\begin{subfigure}[b]{0.485\textwidth}
    \centering\small
    \includegraphics[width=\linewidth]{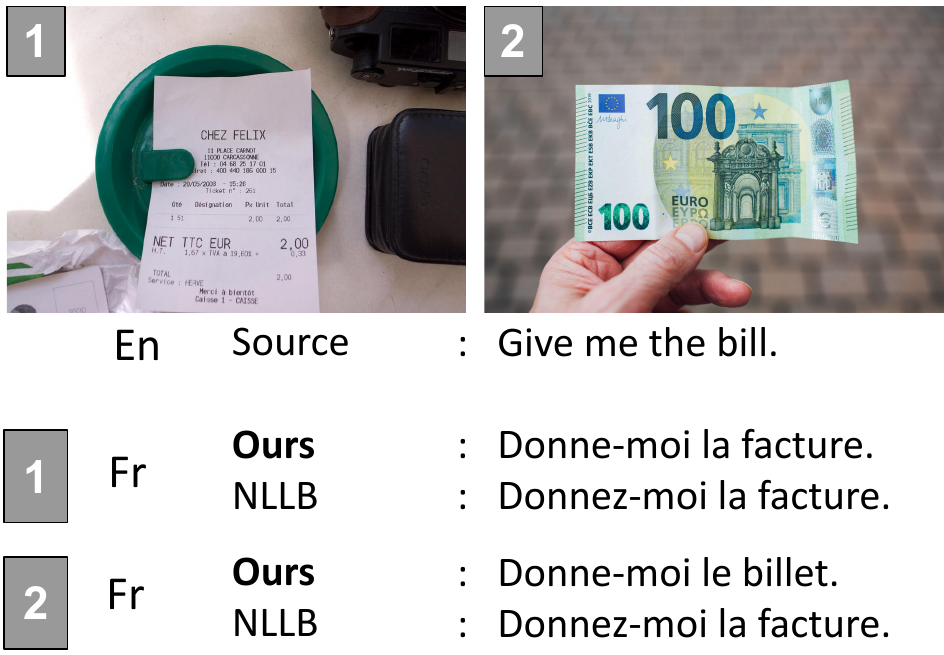}
    \caption{French example from CoMMuTE. The English word `bill' can refer to `paper statement of money owed' or `banknote'. Correctly translated by ZeroMMT in both cases.}
    \label{fig:ex-fr-bill}
\end{subfigure}
\par\bigskip %%%%%%%%%%%%%%%%%%%%%%%%%%%%%%%%%%%%second row
\begin{subfigure}[b]{0.485\textwidth}
    \centering
    \includegraphics[width=\linewidth]{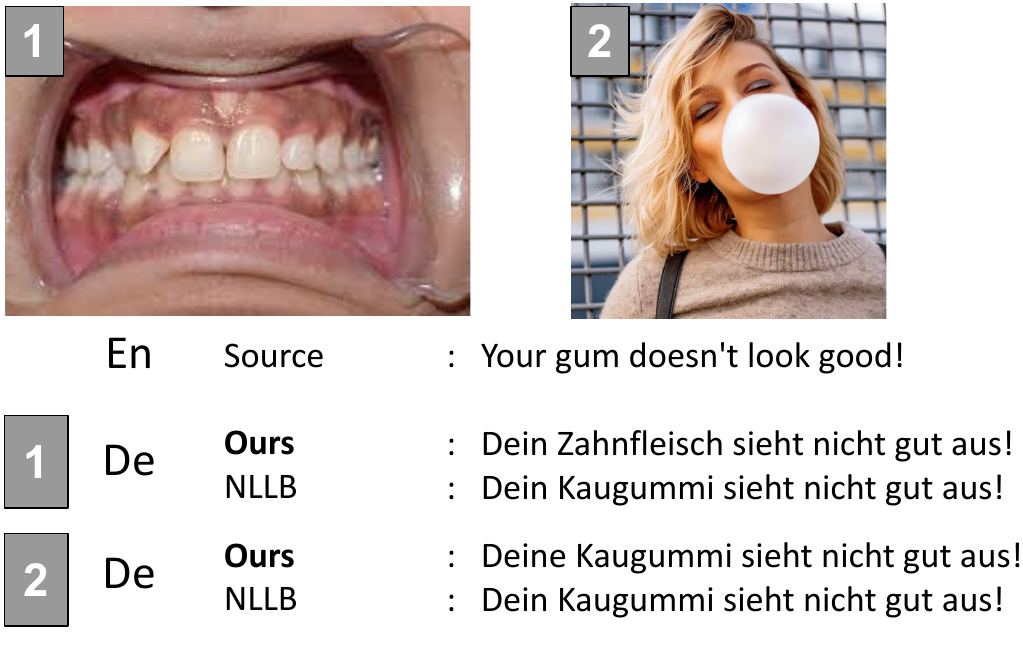}
    \caption{German example from CoMMuTE. The English word `gum' can refer to `mouth tissue' or `chewing gum'. Correctly translated by ZeroMMT in both cases.}
    \label{fig:ex-de-gum}
\end{subfigure}
\hfill
\begin{subfigure}[b]{0.485\textwidth}
    \centering\small
    \includegraphics[width=\linewidth]{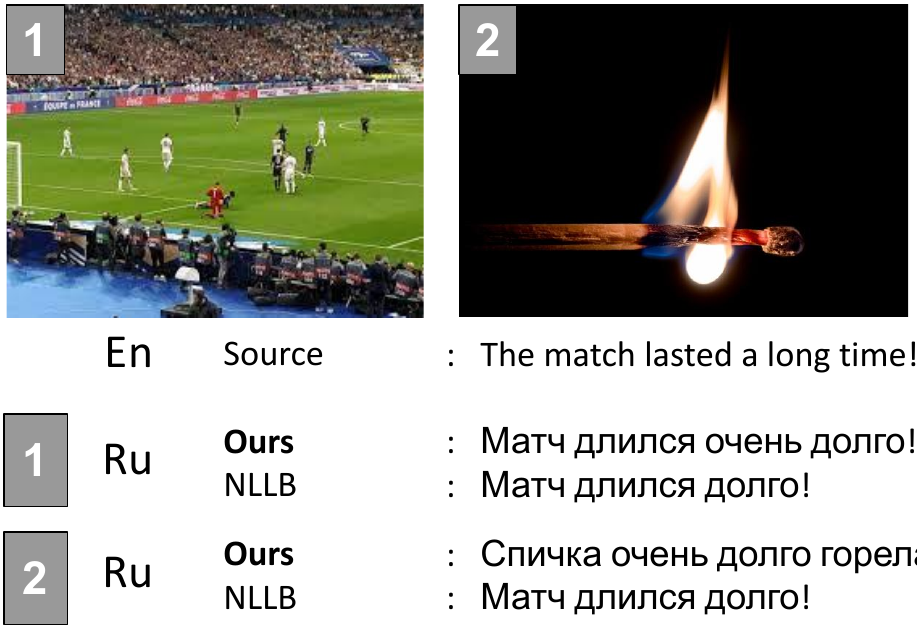}
    \caption{Russian example from CoMMuTE. The English word `match' can refer to `sports game' or `small flammable sticks'. Correctly translated by ZeroMMT in both cases.}
    \label{fig:ex-ru-match}
\end{subfigure}
\par\bigskip %%%%%%%%%%%%%%%%%%%%%%%%%%%%%%%%%%%%second row
\begin{subfigure}[b]{0.485\textwidth}
    \centering
    \includegraphics[width=\linewidth]{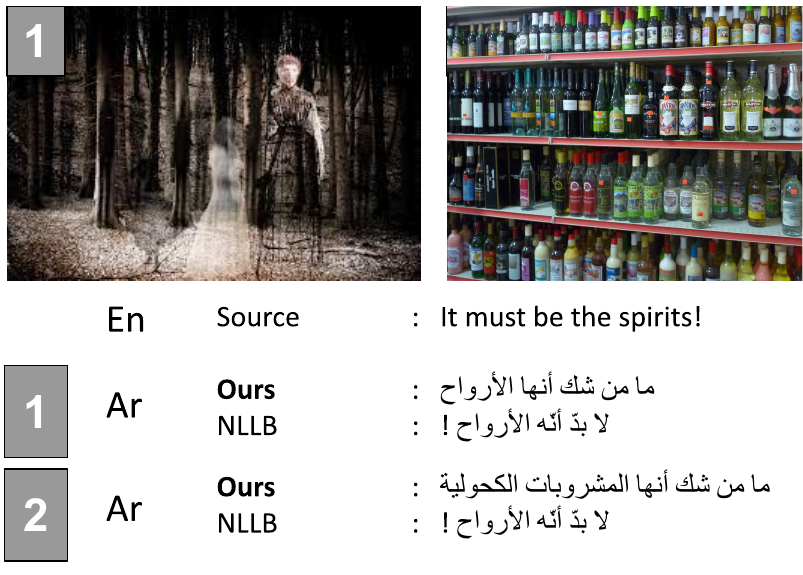}
    \caption{Arabic example from CoMMuTE. The English word `spirits' can refer to `souls' or `alcoholic beverages'. Correctly translated by ZeroMMT in both cases.}
    \label{fig:ex-ar-spirits}
\end{subfigure}
\hfill
\begin{subfigure}[b]{0.485\textwidth}
    \centering\small
    \includegraphics[width=\linewidth]{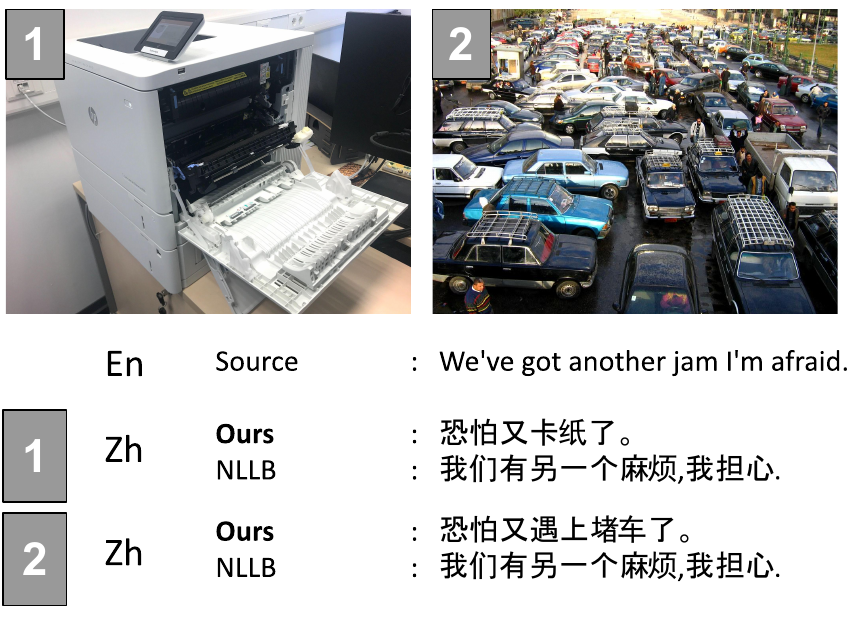}
    \caption{Chinese example from CoMMuTE. The English word `jam' can refer to `paper stuck in a printer' or `being stuck in the traffic'. Correctly translated by ZeroMMT in both cases.}
    \label{fig:ex-zh-jam}
\end{subfigure}
\caption{Translations of CoMMuTE by our approach, ZeroMMT-600M, and the NLLB distilled MT model.}
\end{figure*}

\begin{figure*}[!ht]
    \centering\small
    
\begin{subfigure}[!htbp]{0.46\textwidth}
    \centering
    \includegraphics[width=.9\linewidth]{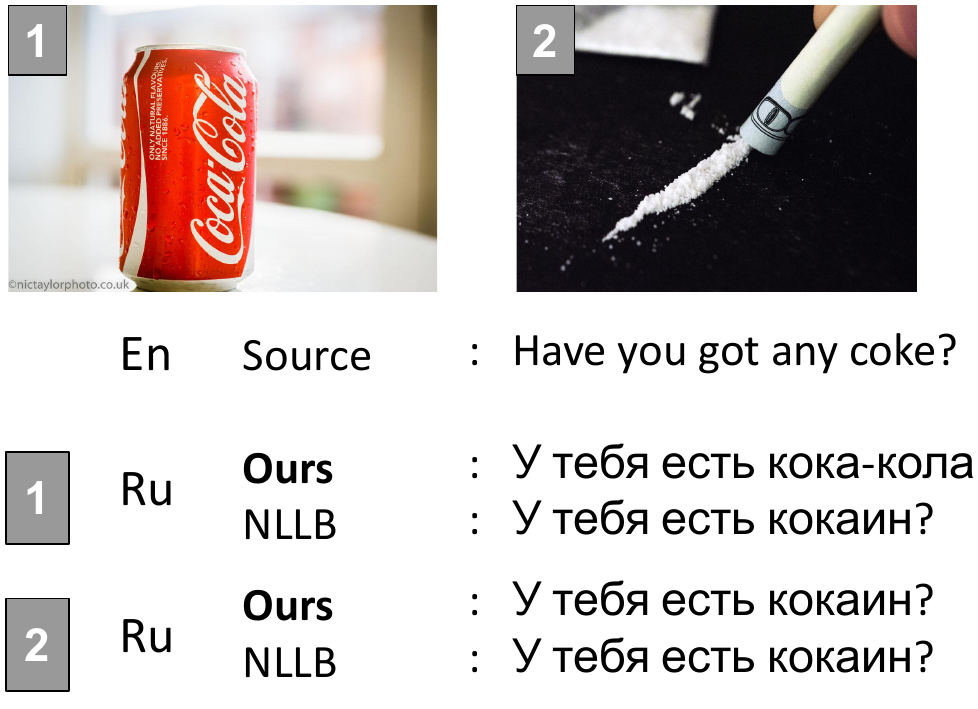}
    \caption{English--Russian example.}
    \label{fig:ru-ex} %\vspace{0.05\linewidth}
\end{subfigure}\vspace{0.05\linewidth}%
\hfill
\begin{subfigure}[!htbp]{0.45\textwidth}
    \centering
    \includegraphics[width=\linewidth]{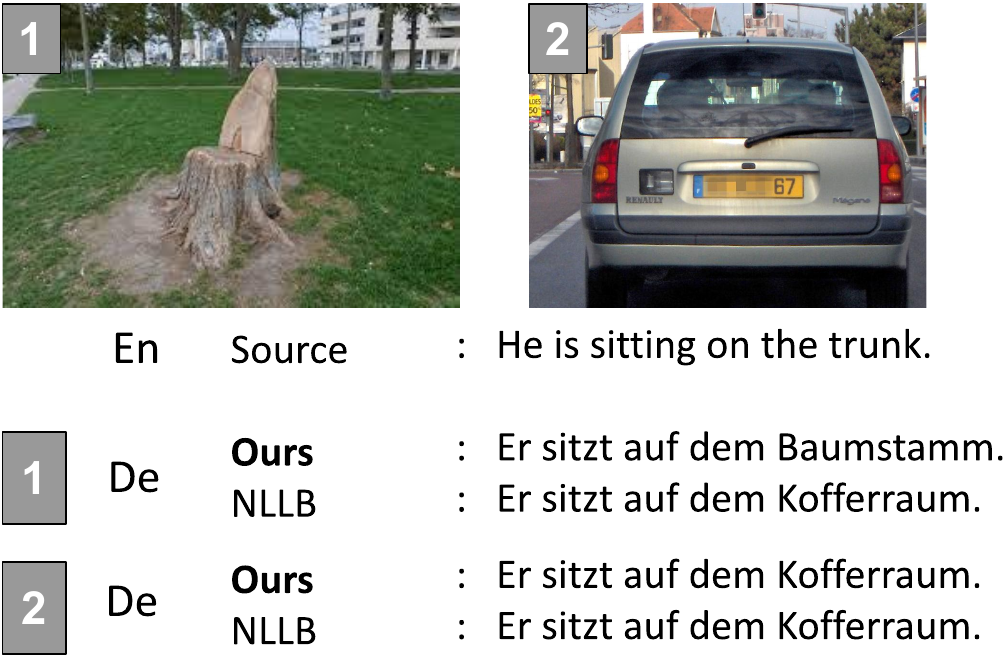}
    \caption{English--German example.}
    \label{fig:de-ex}
\end{subfigure}
\par\bigskip %%%%%%%%%%%%%%%%%%%%%%%%%%%%%%%%%%%%second row
\begin{subfigure}[!htbp]{0.46\textwidth}
    \centering
    \includegraphics[width=.9\linewidth]{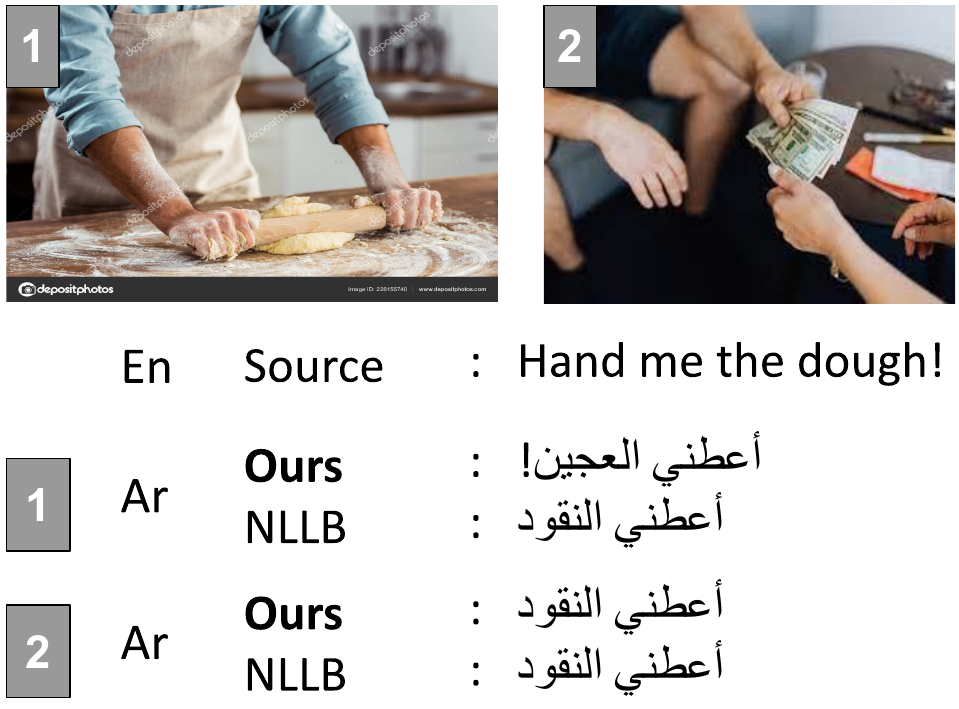}
    \caption{English--Arabic example.}
    \label{fig:arb-ex}
\end{subfigure}%}
\hfill
\begin{subfigure}[!htbp]{0.46\textwidth}
    \centering
    \includegraphics[width=.9\linewidth]{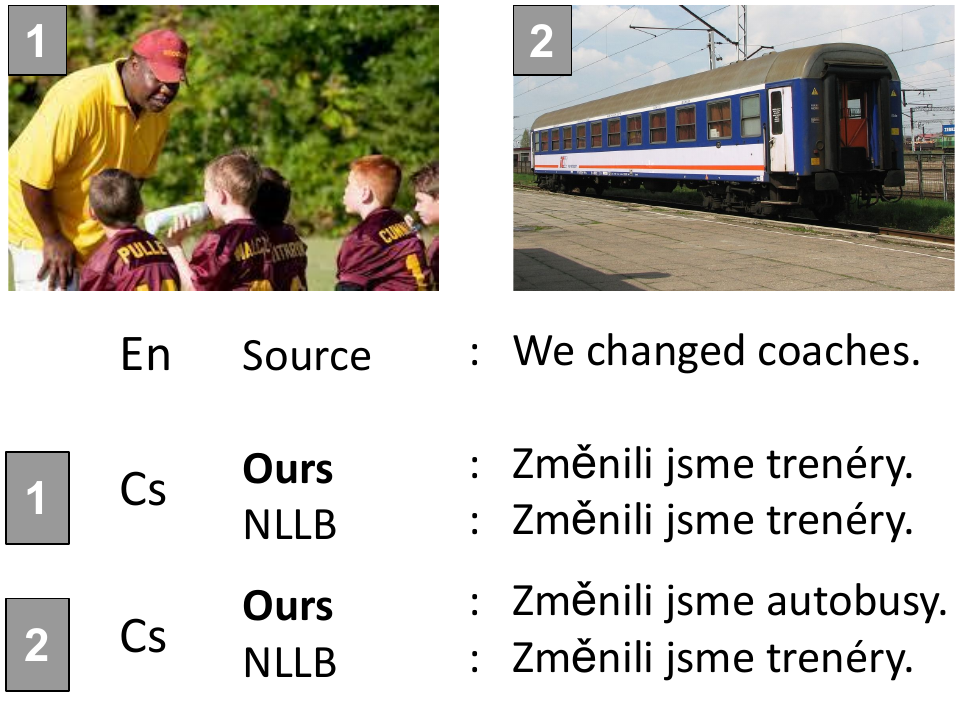}
    \caption{English--Czech example.}
    \label{fig:cs-ex} %\vspace{0.05\linewidth}
\end{subfigure}\vspace{0.05\linewidth}%
\caption{Additional translations of CoMMuTE by our approach, ZeroMMT-600M, and the NLLB distilled MT model.}
\end{figure*}

%\clearpage

\end{document}